\documentclass[sigplan,10pt]{acmart}

\copyrightyear{2018}
\acmYear{2018}
\setcopyright{acmcopyright}
\startPage{1}

\usepackage{booktabs}
\usepackage{subcaption}

\usepackage[draft]{minted} 
                    
\makeatletter
\def\PYG@reset{\let\PYG@it=\relax \let\PYG@bf=\relax%
    \let\PYG@ul=\relax \let\PYG@tc=\relax%
    \let\PYG@bc=\relax \let\PYG@ff=\relax}
\def\PYG@tok#1{\csname PYG@tok@#1\endcsname}
\def\PYG@toks#1+{\ifx\relax#1\empty\else%
    \PYG@tok{#1}\expandafter\PYG@toks\fi}
\def\PYG@do#1{\PYG@bc{\PYG@tc{\PYG@ul{%
    \PYG@it{\PYG@bf{\PYG@ff{#1}}}}}}}
\def\PYG#1#2{\PYG@reset\PYG@toks#1+\relax+\PYG@do{#2}}

\@namedef{PYG@tok@w}{\def\PYG@tc##1{\textcolor[rgb]{0.73,0.73,0.73}{##1}}}
\@namedef{PYG@tok@c}{\let\PYG@it=\textit\def\PYG@tc##1{\textcolor[rgb]{0.24,0.48,0.48}{##1}}}
\@namedef{PYG@tok@cp}{\def\PYG@tc##1{\textcolor[rgb]{0.61,0.40,0.00}{##1}}}
\@namedef{PYG@tok@k}{\let\PYG@bf=\textbf\def\PYG@tc##1{\textcolor[rgb]{0.00,0.50,0.00}{##1}}}
\@namedef{PYG@tok@kp}{\def\PYG@tc##1{\textcolor[rgb]{0.00,0.50,0.00}{##1}}}
\@namedef{PYG@tok@kt}{\def\PYG@tc##1{\textcolor[rgb]{0.69,0.00,0.25}{##1}}}
\@namedef{PYG@tok@o}{\def\PYG@tc##1{\textcolor[rgb]{0.40,0.40,0.40}{##1}}}
\@namedef{PYG@tok@ow}{\let\PYG@bf=\textbf\def\PYG@tc##1{\textcolor[rgb]{0.67,0.13,1.00}{##1}}}
\@namedef{PYG@tok@nb}{\def\PYG@tc##1{\textcolor[rgb]{0.00,0.50,0.00}{##1}}}
\@namedef{PYG@tok@nf}{\def\PYG@tc##1{\textcolor[rgb]{0.00,0.00,1.00}{##1}}}
\@namedef{PYG@tok@nc}{\let\PYG@bf=\textbf\def\PYG@tc##1{\textcolor[rgb]{0.00,0.00,1.00}{##1}}}
\@namedef{PYG@tok@nn}{\let\PYG@bf=\textbf\def\PYG@tc##1{\textcolor[rgb]{0.00,0.00,1.00}{##1}}}
\@namedef{PYG@tok@ne}{\let\PYG@bf=\textbf\def\PYG@tc##1{\textcolor[rgb]{0.80,0.25,0.22}{##1}}}
\@namedef{PYG@tok@nv}{\def\PYG@tc##1{\textcolor[rgb]{0.10,0.09,0.49}{##1}}}
\@namedef{PYG@tok@no}{\def\PYG@tc##1{\textcolor[rgb]{0.53,0.00,0.00}{##1}}}
\@namedef{PYG@tok@nl}{\def\PYG@tc##1{\textcolor[rgb]{0.46,0.46,0.00}{##1}}}
\@namedef{PYG@tok@ni}{\let\PYG@bf=\textbf\def\PYG@tc##1{\textcolor[rgb]{0.44,0.44,0.44}{##1}}}
\@namedef{PYG@tok@na}{\def\PYG@tc##1{\textcolor[rgb]{0.41,0.47,0.13}{##1}}}
\@namedef{PYG@tok@nt}{\let\PYG@bf=\textbf\def\PYG@tc##1{\textcolor[rgb]{0.00,0.50,0.00}{##1}}}
\@namedef{PYG@tok@nd}{\def\PYG@tc##1{\textcolor[rgb]{0.67,0.13,1.00}{##1}}}
\@namedef{PYG@tok@s}{\def\PYG@tc##1{\textcolor[rgb]{0.73,0.13,0.13}{##1}}}
\@namedef{PYG@tok@sd}{\let\PYG@it=\textit\def\PYG@tc##1{\textcolor[rgb]{0.73,0.13,0.13}{##1}}}
\@namedef{PYG@tok@si}{\let\PYG@bf=\textbf\def\PYG@tc##1{\textcolor[rgb]{0.64,0.35,0.47}{##1}}}
\@namedef{PYG@tok@se}{\let\PYG@bf=\textbf\def\PYG@tc##1{\textcolor[rgb]{0.67,0.36,0.12}{##1}}}
\@namedef{PYG@tok@sr}{\def\PYG@tc##1{\textcolor[rgb]{0.64,0.35,0.47}{##1}}}
\@namedef{PYG@tok@ss}{\def\PYG@tc##1{\textcolor[rgb]{0.10,0.09,0.49}{##1}}}
\@namedef{PYG@tok@sx}{\def\PYG@tc##1{\textcolor[rgb]{0.00,0.50,0.00}{##1}}}
\@namedef{PYG@tok@m}{\def\PYG@tc##1{\textcolor[rgb]{0.40,0.40,0.40}{##1}}}
\@namedef{PYG@tok@gh}{\let\PYG@bf=\textbf\def\PYG@tc##1{\textcolor[rgb]{0.00,0.00,0.50}{##1}}}
\@namedef{PYG@tok@gu}{\let\PYG@bf=\textbf\def\PYG@tc##1{\textcolor[rgb]{0.50,0.00,0.50}{##1}}}
\@namedef{PYG@tok@gd}{\def\PYG@tc##1{\textcolor[rgb]{0.63,0.00,0.00}{##1}}}
\@namedef{PYG@tok@gi}{\def\PYG@tc##1{\textcolor[rgb]{0.00,0.52,0.00}{##1}}}
\@namedef{PYG@tok@gr}{\def\PYG@tc##1{\textcolor[rgb]{0.89,0.00,0.00}{##1}}}
\@namedef{PYG@tok@ge}{\let\PYG@it=\textit}
\@namedef{PYG@tok@gs}{\let\PYG@bf=\textbf}
\@namedef{PYG@tok@gp}{\let\PYG@bf=\textbf\def\PYG@tc##1{\textcolor[rgb]{0.00,0.00,0.50}{##1}}}
\@namedef{PYG@tok@go}{\def\PYG@tc##1{\textcolor[rgb]{0.44,0.44,0.44}{##1}}}
\@namedef{PYG@tok@gt}{\def\PYG@tc##1{\textcolor[rgb]{0.00,0.27,0.87}{##1}}}
\@namedef{PYG@tok@err}{\def\PYG@bc##1{{\setlength{\fboxsep}{\string -\fboxrule}\fcolorbox[rgb]{1.00,0.00,0.00}{1,1,1}{\strut ##1}}}}
\@namedef{PYG@tok@kc}{\let\PYG@bf=\textbf\def\PYG@tc##1{\textcolor[rgb]{0.00,0.50,0.00}{##1}}}
\@namedef{PYG@tok@kd}{\let\PYG@bf=\textbf\def\PYG@tc##1{\textcolor[rgb]{0.00,0.50,0.00}{##1}}}
\@namedef{PYG@tok@kn}{\let\PYG@bf=\textbf\def\PYG@tc##1{\textcolor[rgb]{0.00,0.50,0.00}{##1}}}
\@namedef{PYG@tok@kr}{\let\PYG@bf=\textbf\def\PYG@tc##1{\textcolor[rgb]{0.00,0.50,0.00}{##1}}}
\@namedef{PYG@tok@bp}{\def\PYG@tc##1{\textcolor[rgb]{0.00,0.50,0.00}{##1}}}
\@namedef{PYG@tok@fm}{\def\PYG@tc##1{\textcolor[rgb]{0.00,0.00,1.00}{##1}}}
\@namedef{PYG@tok@vc}{\def\PYG@tc##1{\textcolor[rgb]{0.10,0.09,0.49}{##1}}}
\@namedef{PYG@tok@vg}{\def\PYG@tc##1{\textcolor[rgb]{0.10,0.09,0.49}{##1}}}
\@namedef{PYG@tok@vi}{\def\PYG@tc##1{\textcolor[rgb]{0.10,0.09,0.49}{##1}}}
\@namedef{PYG@tok@vm}{\def\PYG@tc##1{\textcolor[rgb]{0.10,0.09,0.49}{##1}}}
\@namedef{PYG@tok@sa}{\def\PYG@tc##1{\textcolor[rgb]{0.73,0.13,0.13}{##1}}}
\@namedef{PYG@tok@sb}{\def\PYG@tc##1{\textcolor[rgb]{0.73,0.13,0.13}{##1}}}
\@namedef{PYG@tok@sc}{\def\PYG@tc##1{\textcolor[rgb]{0.73,0.13,0.13}{##1}}}
\@namedef{PYG@tok@dl}{\def\PYG@tc##1{\textcolor[rgb]{0.73,0.13,0.13}{##1}}}
\@namedef{PYG@tok@s2}{\def\PYG@tc##1{\textcolor[rgb]{0.73,0.13,0.13}{##1}}}
\@namedef{PYG@tok@sh}{\def\PYG@tc##1{\textcolor[rgb]{0.73,0.13,0.13}{##1}}}
\@namedef{PYG@tok@s1}{\def\PYG@tc##1{\textcolor[rgb]{0.73,0.13,0.13}{##1}}}
\@namedef{PYG@tok@mb}{\def\PYG@tc##1{\textcolor[rgb]{0.40,0.40,0.40}{##1}}}
\@namedef{PYG@tok@mf}{\def\PYG@tc##1{\textcolor[rgb]{0.40,0.40,0.40}{##1}}}
\@namedef{PYG@tok@mh}{\def\PYG@tc##1{\textcolor[rgb]{0.40,0.40,0.40}{##1}}}
\@namedef{PYG@tok@mi}{\def\PYG@tc##1{\textcolor[rgb]{0.40,0.40,0.40}{##1}}}
\@namedef{PYG@tok@il}{\def\PYG@tc##1{\textcolor[rgb]{0.40,0.40,0.40}{##1}}}
\@namedef{PYG@tok@mo}{\def\PYG@tc##1{\textcolor[rgb]{0.40,0.40,0.40}{##1}}}
\@namedef{PYG@tok@ch}{\let\PYG@it=\textit\def\PYG@tc##1{\textcolor[rgb]{0.24,0.48,0.48}{##1}}}
\@namedef{PYG@tok@cm}{\let\PYG@it=\textit\def\PYG@tc##1{\textcolor[rgb]{0.24,0.48,0.48}{##1}}}
\@namedef{PYG@tok@cpf}{\let\PYG@it=\textit\def\PYG@tc##1{\textcolor[rgb]{0.24,0.48,0.48}{##1}}}
\@namedef{PYG@tok@c1}{\let\PYG@it=\textit\def\PYG@tc##1{\textcolor[rgb]{0.24,0.48,0.48}{##1}}}
\@namedef{PYG@tok@cs}{\let\PYG@it=\textit\def\PYG@tc##1{\textcolor[rgb]{0.24,0.48,0.48}{##1}}}

\makeatother

\usepackage{listings}
\usepackage{ifthen}

\usepackage[pdf]{graphviz}

\usepackage{hyperref}       
\usepackage{url}            
\usepackage{breakurl}
\usepackage{booktabs}       
\usepackage{nicefrac}       

\usepackage{eqnarray}

\newboolean{showcomments}
\setboolean{showcomments}{false}
\ifthenelse{\boolean{showcomments}}
{ \newcommand{\mynote}[3]{
    \fbox{\bfseries\sffamily\scriptsize#1}
    {\small$\blacktriangleright$\textsf{\emph{\color{#3}{#2}}}$\blacktriangleleft$}}}
{ \newcommand{\mynote}[3]{}}

\newcommand{\loopstack}[0]{LoopStack}
\newcommand{\loopnest}[0]{LoopNest}
\newcommand{\looptool}[0]{LoopTool}

\newcommand{\remove}[1]{}

\newcommand{\dotpath}[1] {#1}

\newcommand{\manualstepsfigpath}[1] {#1}
\newcommand{\versusLLVMfigpath}[1]  {#1}

\begin{document}

\setlength{\pdfpageheight}{\paperheight}
\setlength{\pdfpagewidth}{\paperwidth}

\title{LoopStack: a Lightweight Tensor Algebra Compiler Stack}

\author{Bram Wasti}
\authornote{Corresponding authors: \{bwasti,zlateski\} at fb dot com.}
\affiliation{ \institution{Meta AI} }

\author{José Pablo Cambronero}
\authornote{Author contributed while interning with Meta and completing studies at MIT.}
\affiliation{ \institution{MIT, Meta AI} }

\author{Benoit Steiner}
\affiliation{ \institution{Meta AI} }

\author{Hugh Leather}
\affiliation{ \institution{Meta AI} }

\author{Aleksandar Zlateski}
\authornotemark[1]
\affiliation{ \institution{Meta AI} }








\settopmatter{printacmref=false} 
\renewcommand\footnotetextcopyrightpermission[1]{} 
\pagestyle{plain} 

\date{}
\thispagestyle{empty}

\begin{abstract}
  We present \loopstack{}, a domain specific compiler stack for tensor
operations, composed of a frontend, \looptool{}, and an efficient
optimizing code generator, \loopnest{}. This stack enables us to
compile entire neural networks and generate code targeting the AVX2,
AVX512, NEON, and NEONfp16 instruction sets while incorporating
optimizations often missing from other machine learning compiler
backends.

We evaluate our stack on a collection of full neural networks and
commonly used network blocks as well as individual operators, and show
that \loopstack{} generates machine code that matches and frequently
exceeds the performance of in state-of-the-art machine learning
frameworks in both cases.  We also show that for a large collection of
schedules \loopnest{}'s compilation is orders of magnitude faster than
LLVM, while resulting in equal or improved run time performance.
Additionally, \loopstack{} has a very small memory footprint -- a
binary size of 245KB, and under 30K lines of \emph{effective} code
makes it ideal for use on mobile and embedded devices.

\end{abstract}

\maketitle


\section{Introduction}

The availability of massive amounts of computing power has fueled the
explosive growth of machine learning techniques for almost a decade
and has greatly affected the design of modern hardware. Companies,
such as NVIDIA with their tensor cores, Intel/AMD with their
specialized AVX, FMA, and VNNI instructions, ARM with the Neon and SVE
extensions to their instruction set, and Apple with their M1 CPU and
its matrix coprocessor, have tailored the computational capabilities
of their respective hardware to better serve the needs of deep
learning workloads.

However, leveraging the raw computational power of such hardware for
the purpose of fast execution of machine learning models remains a
challenge. Several approaches have been proposed over time, but they
all suffer from significant shortcomings.

The first approach is to rely on libraries of optimized primitive
tensor operations, such as cuDNN~\cite{chetlur2014cudnn},
OneDNN~\cite{onednn}, or XNNPACK~\cite{xnnpack}. Unfortunately this
approach has three main downsides. First, to cover an ever growing set
of use cases, these libraries tend to be large in codesize, which
hinders their utilization on many devices, such as smartphones or IoT
devices. Second, their development and maintenance require large
amounts of manpower. Third, operators can only exchange data through
global memory, which can be a significant bottleneck, especially in
the case of low arithmetic intensity operators such as the activation
functions.

To avoid these problems, another approach pioneered by projects such
as Halide~\cite{ragan2013halide} and TVM~\cite{tvm} represent tensor
computations, such as the ones underlying deep learning, using a
declarative domain specific language. This high-level representation
is then scheduled and lowered into LLVM intermediate representation,
and then compiled into instructions that can be executed directly on
hardware by the LLVM compiler. However this approach suffers from very
large compilation times, thus making certain techniques, such as fast
auto-tuning impractical.

To overcome these limitations, we introduce LoopStack, a lightweight
toolchain designed specifically for deep learning workloads. To
express computations, LoopStack introduces a \textbf{domain specific
  representation} based on Einstein's notation. This representation
can capture many common dense neural network operations in a concise
form. LoopStack also provides a \textbf{frontend} capable of
converting neural networks, expressed as dataflow graphs, into our
representation.  Finally, we present a \textbf{code generator} that
can extremely quickly convert this representation into highly
optimized code for X86 and ARM CPUs.

Unlike other approaches that rely on off-the-shelf components such as
LLVM, \loopstack{} was designed from the ground up to target machine
learning workloads. This enables our system to make the following
contributions:

\begin{itemize}
\item Milliseconds compile times, a fraction of time compared to
  traditional compilers.
\item Performance comparable to or exceeding that of hand--optimized
  libraries such as MKL--DNN or XNNPACK.
\item Small code footprint resulting in a lighweight binary.
\end{itemize}


\section{Tensor Contractions}

Tensor operations, such as inner, outer, and cross product, as well as
matrix multiplication, trace, transpose, and many other tensor
operations can be expressed as special cases of contraction. In fact,
with a few extensions, contractions can encode all the operations that
the most popular deep neural networks rely on. We demonstrate how to
express computations over tensors as generalized contractions. We also
discuss the challenges inherent in optimizing (aka scheduling) these
contractions.

\subsection{Generalized Tensor Contractions}

Before introducing generalized, $d$-dimensional tensor contractions,
we first present a simple example in $2$-dimensions -- the matrix
multiplication operation.  Figure~\ref{fig:mm-background-example}
demonstrates two different, though equivalent, ways that describe the
matrix multiplication operation.  The imperative pseudocode in
Figure~\ref{mm-imperative} shows the typical set of 3 nested for-loops
iterating over dimensions $i$ and $j$ of the output matrix ($C$) and
the \emph{reduction dimension} $k$.  Note the reduction over $k$ is
explicit in this notation.

In the tensor index (Einstein) notation~\cite{ricci1900methodes,
  einstein1923grundlage, kjolstad2017tensor}
(Figure~\ref{mm-einstein}), the $i$ and $j$ index variables appear on
both the left-- and right--hand side of the formula indicating that in
order to compute the corresponding entry in $C$ we index $A$ and $B$
with the same values of $i$ and $j$. The index variable $k$,
represents a reduction variable, as it appears solely on the
right--hand side the formula. In particular, the for--loop over
varying values of $k$ for a specific assignment to $i$ and $j$, and
the corresponding aggregation (i.e. contraction), is left implicit.
Similarly, the for--loops over varying values of $i$ and $j$, needed
to populate $C$, are left implicit.

We use generic $\oplus$ and $\otimes$ notation to emphasize that
different operators can be used to instantiate a generalized matrix
\emph{multiplication} operation.

\begin{figure}[h]
  \begin{subfigure}{.45\textwidth}
    \begin{lstlisting}[basicstyle=\ttfamily\scriptsize, language=python, breaklines=true]
for i:
    for j:
        C[i, j] = alpha * C[i, j];
        for k:
            C[i, j] += beta * A[i, k] * B[k, j]
    \end{lstlisting}
    \caption{Pseudocode for basic imperative matrix-multiplication.}
    \label{mm-imperative}
  \end{subfigure}
  \begin{subfigure}{.45\textwidth}
    \begin{gather*}
      C_{i,j} = \alpha C^{in}_{i, j} \oplus \beta(A_{i, k} \otimes B_{k, j})
    \end{gather*}
    \caption{Equivalent declarative Einstein notation for matrix multiplication, with  $\oplus=+$ and $\otimes=*$.}
    \label{mm-einstein}
  \end{subfigure}
  \caption{Matrix multiplication}
  \label{fig:mm-background-example}
\end{figure}

We can generalize the matrix-multiplication example to
$d$--dimensions, where our inputs are now $d$--dimensional tensors, to
result in \emph{tensor contractions}.  Following notation presented
in~\cite{gareev2018high}, let $A$, $B$, and $C$ now be d-dimensional
tensors (for potentially different $d$). Let the set of dimensions for
$A$, $B$, and $C$ be denoted by $I_A$, $I_B$, and $I_C$, respectively.
Let $I_{AC}$ be the dimensions of $C$ present in $A$, while $I_{AR}$
are reduction dimensions\footnote{Dimensions contracted by a
user-defined operation, such as the shared dimension of the input
matrices in a matrix multiplication.} in $A$.  Similarly $I_{BC}$ are
dimensions of $C$ present in $B$, while $I_{BR}$ are reduction
dimensions in $B$.  We now write:
\[
    C_{I_C} = \alpha C^{in}_{I_C} \oplus \beta (A_{I_{AC} \cup I_{AR}} \otimes B_{I_{BC} \cup I_{BR}})
\]
We further extend this notation to allow indexing into a tensor to be
an affine transformation of the indices.

Finally, we extend our notation to allow for an element-wise operation
that initializes the output tensor (\emph{pre-operation}) and an
element-wise operation that can transform a value before final storage
into the output tensor (\emph{post-operation}). We can now write
\[
    C_{(I_C)} = \text{post}(\text{pre}(C^{in}_{f_C(I_C)}) \oplus \beta (A_{f_A(I_{AC} \cup I_{AR})} \otimes B_{f_B(I_{BC} \cup I_{BR})}))
\]
where $f_C$, $f_A$ and $f_B$ are affine transformations and pre/post
are the element-wise pre- and post- operations.  These concepts are
useful for expressing biases and activation functions in machine
learning workloads.

With these extensions, we can not only express GEMM, GEMV, and GEVM
operations, but many more operations as well. Such as:
\begin{itemize}
\item Convolutions -- $O_{r, c} = I_{r + k, c + j} \cdot  \omega_{k, j}$
\item pooling -- $O_{r, c} = \max(I_{2r + k, 2c + k})$
\item reductions -- $O_{r} = I_{r, c}$
\item transpositions -- $O_{r, c} = I_{c, r}$
\item concatenations -- $O_{r, c'} = I^1_{r, c'}$; $O_{r, |c'| + c''} = I^2_{r, c''}$
\item broadcast -- $O_{r, c} = I_{r}$
\end{itemize}

\subsection{Optimizing Generalized Contractions}

The way we carry out these generalized contraction computations has a
significant impact on performance. For example, a naive 3-nested
for-loop implementation of matrix multiplication of
figure~\ref{mm-imperative} will result in long running times for
non-trivially sized matrices. In contrast, efficient implementations,
as proposed by many researchers over the past few
decades~\cite{goto2008anatomy, park2000efficient, smith2014anatomy,
  lam1991cache, gunnels2001family, jia2018optimizing,
  zlateski2019anatomy}, might modify the imperative loops to tile
(i.e. split into sub-matrices of appropriate sizes) the input matrices
to align with architecture-dependent resources such as cache sizes,
re-order loop dimensions to re-use data in the innermost loop, exploit
architecture features such as vector instructions to operate on
multiple elements at a time, and emit extended instructions such as
fused-multiply-add, which combine multiplication and accumulation at
the instruction level. Further, the data might be kept in
\emph{exotic} memory layouts, such as the channel--interleaved format
often used on Intel machines~\cite{onednn, zlateski2017compile}, or
the matrix tile interleaved format proposed by Jia et
al.~\cite{jia2018optimizing, zlateski2019anatomy}.

Typical scheduling operations include: re-ordering, splitting, fusing,
parallelizing, vectorizing, and unrolling loops
~\cite{ragan2013halide, grosser2011polly}.  The scheduling options for
a single task such as matrix-multiplication highlight the
combinatorial challenge underlying scheduling. Different computations
and platforms require varying schedules to deliver performance gains.
This results in a challenging optimization problem. Historically,
varying techniques have been put forward to tackle this optimization,
ranging from expert-based manual optimization, analysis-driven
heuristic optimization ~\cite{grosser2011polly, leben2019polyhedral},
to recent advances in data-driven automated schedule
exploration~\cite{tvm, adams2019learning}

In the remainder of this paper, we introduce both the frontend
(\looptool{}) and backend (\loopnest{}) of \loopstack{} as well as a
TUI for manual tuning and a script for automatic tuning built on top
of \loopstack{}.


\section{\loopstack{} Design}

\loopstack{} is composed of a frontend (\looptool{}) and a backend
(\loopnest{}).  This design separates the representation of the
program being described from the logic to emit target-specific code.

The frontend (\looptool{}) is designed to enable expression of both
the user's mathematical intent (a chain of extended tensor
contractions) as well as the preferred execution of the program -- the
order in which basic mathematical operations should be executed
(i.e. the schedule) to perform the desired tensor contraction.  This
is done by defining a structured intermediate representation composed
exclusively of a dataflow graph (DFG) as well as a loop order
annotation language that denotes how each node is executed.  The
omission of control flow in the representation of the program is
consistent with a large majority of neural network, which can be well
represented as purely functional operations on N-dimensional tensors.
The design of this IR -- the design of our DFG, as well as the loop
order annotations, is explained thoroughly in section
\ref{loop-tool-core}.

The backend, \loopnest{}, is designed to generate optimized
machine--code efficiently that strictly adheres to the user's input.
\loopnest{} exposes an API that allows the user to define a specific
loop nest order in which the computation should be performed;
\loopnest{} will not make any attempt to change the provided nest
order or use any logic to recover user intent.  In short --
\loopnest{} will generate machine code that performs the exact
computation that the user requested, and in the same order.  However
\loopnest{} will attempt to apply all known low level optimizations
that are relevant to the target hardware, such as loop
unrolling\cite{jiang2018efficient,zheng2018optimizing,zhao2018hardware,wiki:Loop_unrolling,kennedy2001optimizing,jeffers2013intel,spampinato2014basic},
vectorization~\cite{zheng2018optimizing,kennedy2001optimizing,cornea2015intel,jeffers2013intel,spampinato2014basic},
reordering of independent
instructions~\cite{jiang2018efficient,zhao2018hardware,wiki:Instruction_pipelining,kennedy2001optimizing,cornea2015intel,jeffers2013intel,spampinato2014basic},
software
pipelining~\cite{zhao2018hardware,wiki:Instruction_pipelining,kennedy2001optimizing,cornea2015intel,jeffers2013intel,spampinato2014basic},
minimizing the amount of auxiliary operations.  More about \loopnest{}
code generation and optimizations is discussed in section
\ref{loop-nest-core}.

\subsection{Design principles and goals}

\subsubsection{Simplicity}

The goal of simplicity extends to both the description of intended
mathematical operations and the language of defining the execution
order (the schedule) of the resultant loops.

We use an Einstein-like notation~\cite{einstein1923grundlage} embedded
in Python, described in~\ref{loop-tool-core}, as it allows the user to
easily describe the intended mathematical operations even for complex
machine learning models.  For instance, defining a multi layer
perceptron (MLP), which is often used for NLP~\cite{},
reccomendation~\cite{} models, etc.., can be achieved as seen in in
the Listing~\ref{einsum-like}.

\begin{listing}[H]
\remove{
\begin{minted}[fontsize=\small]{python}
    HL1[b, i] += W1[i, j] * Input[b, j]
    Act_1[b, i] = lt.relu(HL1[b, i])
    HL2[b, k] += W2[k, i] * HL1[b, i]
    # ...
    Out[b] = lt.sigmoid(WN[o, p] * HLN[b, p])
\end{minted}
}
\input{listing1.pygtex}
\caption{MLP defined with our Einstein notation-like syntax}
\label{einsum-like}
\end{listing}

We provide a minimal IR and limit the API such that almost all
optimization operations can be decomposed into a series of operations
on individual nodes in the DFG.  Wrappers are provided to manipulate
nodes in bulk (e.g. to change the loop order of both the
multiplication and the addition in a matrix multiplication at once),
yielding an interface similar to Halide's pipeline scheduling
states\cite{adams2019learning}. However, unlike in Halide, it is
impossible to represent illegal schedules in our IR.  We discuss the
optimizations in detail in section \ref{optimization}.

There are only a handful of global optimization parameters, such as
the unroll limit and the parameters describing the target hardware,
most of which are encoded inside the \loopnest{}.

\subsubsection{Performance Predictability \& Fast Feedback}

Traditional compilers perform expensive analysis in order to
understand user intent and optimize execution, while performing
equivalent computation. This approach has two major drawbacks.  First,
the user does not get accurate feedback about the quality of their
intended schedule of execution; and second, the compilation time is
significantly increased due to expensive analysis.  \loopstack{} takes
a different approach.  Once the user annotates the IR, no layer in
\loopstack{} attempts to understand the user's intent and/or reorder
the intended schedule provided by the user, except in very few,
hardware--specific cases.  Currently, these cases only include
reordering mathematical operations inside the innermost unrolled loop.
Such optimizations have relatively minor effect on the overall
performances -- they increase the utilization of the underlying
hardware by a few percentage points, but also allow the user to
squeeze the last bits of the available computational power.  In
addition, such optimizations highly depend on the limitations of the
target hardware, and are not easy to expose to the user.

This approach allows \loopstack{} to perform extremely fast
compilation (on the order of milliseconds), as well as allow the user
to understand the quality of their intended schedules and rapidly
explore scheduling ideas.  In addition, this approach can greatly
benefit autotuning~\cite{adams2019learning,zheng2020ansor} algorithms
by leveraging both the predictability and efficiency of benchmarking
schedules.


\section{Frontend (\looptool{})}
\label{loop-tool-core}

In order to allow users to effortlessly describe the desired
computation of neural network workloads using \loopstack{}, we built a
domain specific language (DSL) embedded in Python and a compiler
front-end that we termed \looptool{}. \looptool{} exposes a
declarative API for user definition of computation and operates on an
IR composed of an annotated dataflow graph (DFG) consisting of
N-dimensional tensor operations.  The DFG describes the underlying
computation, memory layouts and execution.  \looptool{} then lowers
the DFG into a series of loops, which are then compiled with
\loopnest{}.

\subsection{Declarative API}

\begin{figure}
\begin{lstlisting}[basicstyle=\ttfamily\scriptsize, language=python, breaklines=true]
import loop_tool as lt
M, N, K = lt.Var("m"), lt.Var("n"), lt.Var("k")

A = lt.Tensor([M, K])
B = lt.Tensor([K])
C = lt.Tensor()

C[m, n] += A[m, k] * B[k]
\end{lstlisting}
\caption{\looptool{}'s Python embedded declarative DSL}
\label{fig:looptool-lang-compute}
\end{figure}

Figure~\ref{fig:looptool-lang-compute} shows an example of matrix
multiplication in \looptool{}'s declarative Python DSL. The expression
\lstinline{lt.Var("m")} defines an indexing variable with the
corresponding name. Next, the expression \lstinline{lt.Tensor([M, K])}
defines a 2-dimensional tensor. Note that \looptool{} uses symbolic
(i.e. named) dimensions~\cite{namedtensordims}, which simplify
indexing semantics and encourages a simple interaction model for
manipulating traversal and memory layouts of higher dimensional
tensors.  The expression \lstinline{C[m, n] += A[m, k] * B[k]} defines
computation using the aforementioned tensor (Einstein) notation; here
\lstinline{k} is a reduction dimension.  \looptool{}'s computation
language supports element-wise computations (with broadcast
semantics), associative reductions across arbitrary dimensions, and a
restricted set of indexing semantics (see Section~\ref{view-nodes}).

\subsection{Intermediate Representation (IR)}

\begin{figure}[!htb]
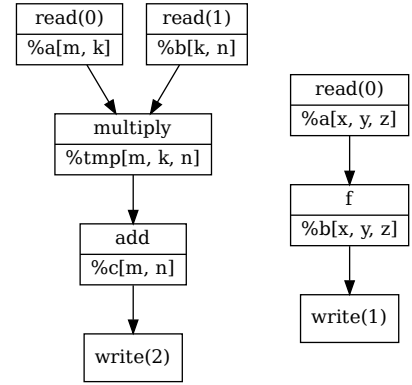
\centering
\begin{minipage}{0.42\columnwidth}
    \digraph[scale=0.5]{\dotpath{matmulir}}{
      read_a -> mul;
      read_b -> mul;
      mul -> add -> write_c;
      read_a[shape=record,label="{ read(0) | \%a[m, k] }"];
      read_b[shape=record,label="{ read(1) | \%b[k, n] }"];
      mul[shape=record,label="{ multiply | \%tmp[m, k, n] }"];
      add[shape=record,label="{ add | \%c[m, n] }"];
      write_c[shape=record,label="write(2)"];
    }
\end{minipage}
\begin{minipage}{0.42\columnwidth}
\digraph[scale=0.5]{\dotpath{examplelowering}}{
  read -> f -> write;
  read[shape=record,label="{ read(0) | \%a[x, y, z] }"];
  f[shape=record,label="{ f | \%b[x, y, z] }"];
  write[shape=record,label="write(1)"];
}
\end{minipage}

\caption{Matrix multiplication in \looptool{} (left).  A point-wise
  application of the function $f$ across all three dimensions of
  \texttt{\%a} (right).}
\label{fig:mm-ir}
\end{figure}

\looptool{}'s DFG is based on an intermediate representation with
annotations on each node.  Each node is associated with its output --
a virtual buffer that is materialized according to the user provided
schedule. Figure~\ref{fig:mm-ir} (left) demonstrates a matrix
multiplication without annotations.  A node's output size is not
materialized until the IR is lowered to loops. The materialization
logic always attempts to minimize the total memory used. For example,
two nodes operating on virtual buffer of size $N$ in a shared loop
over $N$ do not need to allocate the full $N$ elements of memory for
their intermediate. Instead, the intermediate will be of size 1 and
reused $N$ times.  This can be seen in Figure~\ref{lower1}.

\looptool{}'s DFG has three fundamental types of nodes.

\subsubsection{Read/write nodes}

Read and write nodes denote reads or writes from and into user
provided memory, as well as associated layouts and sizes.  These nodes
contain an ordered list of symbolic dimensions.  The ordered list of
dimensions represents a row-major order (lexographical order) of
either input or output memory.  Because \loopstack{} is embedded in
either Python or C++ applications, these nodes are used to interface
with other operations that are not handled by \loopstack{}.  An
example would be the specification of NCHW\cite{pytorch} or
NHWC\cite{abadi2016tensorflow} layouts for the inputs to convolution
operations, both of which are well handled and trivially manipulated
in the \looptool{} IR.  In the IR, read nodes take no inputs whereas
write nodes take a single input and have no outputs.  These are
fundamentally source and sink operations in the graph.

\subsubsection{Arithmetic nodes}

Arithmetic nodes operate on virtual buffer and output a single virtual
buffer.  The ordered list of dimensions associated with these nodes
denotes how output memory should be laid out (given the corresponding
scope of their execution).

Arithmetic nodes, unlike read and write nodes, can take multiple
inputs.  This represents the typical operation applied to the inputs.
For example, addition of two inputs works as expected to yield a
single output.

Extending this concept to higher dimensions forces us to consider when
two inputs do not have the same dimensions.  Due to the symbolic
nature of the dimensions in \looptool{}, we can distinguish two
dimensions of the same size as being mathematically distinct.  To
handle application of arithmetic in higher dimensions we employ
implicit broadcasting akin to Numpy\cite{oliphant2006guide} semantics.
Dimensions not present in the output are implicitly reduced over
according to the arithmetic of the associated node.

\subsubsection{View nodes}

\label{view-nodes}
The final type of node admits the representation of symbolic indexing
constraints over tensor dimensions, allowing powerful view semantics.
Any affine combination of iteration over the output dimensions can be
used to index into an input dimension.  This enables us to represent
windowed operations as well as concatenations.  By using index
constraints rather than index equations, we preserve the meaning of
the underlying computation and can freely split and permute variables
and layouts.  Further, by keeping indexing math symbolic, we can still
apply the scope based memory minimization logic.

\subsection{Lowering to loops}

\label{lower-to-loops}
\looptool{} lowers its IR to an internal loop tree structure before
invoking \loopnest{}.  This is done by traversing the DFG
topologically and eagerly emitting loops that are required for each
node. A reference to the current innermost loop is maintained
throughout this entire process, with each subsequent loop nesting
inside the reference.  If any node requires a loop that is an ancestor
to the current reference, there is no need to emit the loop again (it
would be incorrect to do so) and it is skipped.  This naturally
induces loop coalescing (typically referred to as loop fusion).

Consider nodes shown in Figure~\ref{fig:mm-ir} (right). We can assume
an annotation of \texttt{\%a} and \texttt{\%b} both with loop order
\texttt{[x, y, z]}. When visiting \texttt{\%a}'s node, we emit the
loop tree shown in Listing~\ref{lower0}.

Later, while visiting \texttt{\%b}'s node, we note that the reference
(which is at location \texttt{\%a}) contains a loop for \texttt{x},
\texttt{y}, and \texttt{z} in order.  We thus reuse all loops and
implicitly "fuse" node \texttt{\%b}.  This is shown in
Listing~\ref{lower1}

We find the nodes can be executed in the same innermost loop.  The
resultant allocation size of \texttt{\%a} would be $1$ in this case.

\begin{listing}[H]
\remove{
\begin{minted}[fontsize=\small]{python}
iter x:
  iter y:
    iter z:
      %a[x, y, z] = read(x, y, z)
      ...
\end{minted}
}
\input{listing2.pygtex}
\caption{Lowering \texttt{\%a} emits three new loops.}
\label{lower0}
\end{listing}

\begin{listing}[H]
  \remove{
    \begin{minted}[fontsize=\small]{python}
iter x:
  iter y:
    iter z:
      %a[x, y, z] = read(x, y, z)
      %b[x, y, z] = f(%a[x, y, z])
\end{minted}
}
\input{listing3.pygtex}

\caption{Lowering \texttt{\%b} reuses all loops.}
\label{lower1}
\end{listing}

However, if the nodes had different loop annotations, such as
\texttt{\%b} annotated with \texttt{[x, z, y]}, we would not be able
to share every loop. The memory allocated for \texttt{\%a} in that
case would be of size $|y| \cdot |z|$ This is shown in
Listing~\ref{lower2}.

\begin{listing}[H]
  \remove
  { \begin{minted}[fontsize=\small]{python}
iter x:
  iter y:
    iter z:
      %a[x, y, z] = read(x, y, z)
  iter z:
    iter y:
      %b[x, y, z] = f(%a[x, y, z])
    \end{minted}
  }
  \input{listing4.pygtex}

\caption{Only loop \texttt{x} is shared across the two nodes.}
\label{lower2}
\end{listing}

In the case of reductions, we cannot always share loops.  Consider a
reduction node \texttt{\%R} over variable \texttt{z} with loop order
\texttt{[x, y, z]} depended on by by \texttt{\%a} with the same loop
order; see Listing~\ref{lower3}. The loop for \texttt{z} is required
to run twice for correctness.  This type of behavior, while necessary
for reductions, may be preferable in other contexts as well.

\begin{listing}[H]
  \remove
  { \begin{minted}[fontsize=\small]{python}
iter x:
  iter y:
    iter z:
      %R[x, y] = reduction(...)
    iter z:
      %a[x, y, z] = %R[x, y] + ...
\end{minted}
}
\input{listing5.pygtex}

\caption{Sharing loop \texttt{z} between \texttt{\%R} and \texttt{\%a} would be mathematically incorrect.}
\label{lower3}
\end{listing}

Manually preventing loop sharing can induce larger intermediate memory
allocations.  This is often beneficial when computation benefits from
packing memory into cache--friendly layout before computation.  To
express this, \looptool{} has a second form of annotation for nodes
called \textit{staging} that prevents reuse of specific loops.  In
listing \ref{lower1}, \texttt{\%a} and \texttt{\%b} share an entire
loop nest.  If we were to \textit{stage} the loop for \texttt{\%a}
over \texttt{z}, the resultant lowering would increase the allocation
size of \texttt{\%a} to $|z|$ (Listing~\ref{lower4}).

\begin{listing}[H]
  \remove
  { \begin{minted}[fontsize=\small]{python}
iter x:
  iter y:
    iter z:
      %a[x, y, z] = read(x, y, z)
    iter z:
      %b[x, y, z] = f(%a[x, y, z])
\end{minted}
}
\input{listing6.pygtex}
\caption{\texttt{z} is staged, so \texttt{\%a} is materialized with an
  allocation of size $|z|$.}
\label{lower4}
\end{listing}


\section{Backend (\loopnest{})}
\label{loop-nest-core}

\loopnest{} is a domain specific compiler for a series of nested
loops, with user specified innermost operation.  \loopnest{} is
designed to have extremely short compilation times and use well known
and studied high-performance computing (HPC) optimizations to generate
high performance code.

\subsection{\loopnest{}'s HPC Philosophy}

While traditional compilers typically perform multiple optimization
passes, some of which might be repeated, \loopnest{}'s is designed to
only do a limited number of optimizations, but do them very well.
These include HPC optimizations that are commonly used in expert
designed, custom assembly, or code generator primitives for specific
problems, such as matrix multiplications~\cite{van2015blis,
  barrachina2008evaluation, wang2014intel, heinecke2016libxsmm,
  heinecke2015libxsmm, khudia2018open} and other primitives used in
machine learning~\cite{chetlur2014cudnn, zlateski2018deeper,
  zlateski2019anatomy, jia2018optimizing, onednn, elsen2020fast,
  heinecke2016high}.  Additional HPC optimization might include
instruction reordering, r--sum, and other optimizations suggested by
optimization manuals for the target hardware~\cite{intel2014intel,
  arm2016arm}.

\loopnest{}'s generated code directly reflects the computation
requested by the user: in particular, operations are performed in
user-specified order.  Generating code directly from the user-defined
execution order simplifies code generation logic, which results in a
more intuitive mapping between the quality of the provided execution
order and the observed performance of the code.  In addition, the
provided (high level) information about the loop order and sizes are
used for \loopnest{}'s simple register allocation approach, described
below.

While having default choices based on the target hardware manuals,
\loopnest{} optionally delegates the choice of which loops to be
unrolled.  The user may override the hardware tuned default maximum
number of unrolled instructions, and \loopnest{} will determine the
outermost loop such that the number of unrolled instructions doesn't
exceed the user specified value.

\subsubsection*{Vectorization}

\loopnest{} assumes that the user intent is to vectorize the innermost
loop in the user provided schedule.  This design simplifies the logic,
while not introducing any limitations to the user -- the elements of
the vector operation are executed at the same time, thus naturally
belong to the innermost loop.  \loopnest{} will, thus, attempt to
vectorize the innermost loop. However, in certain scenarios, when the
data accessed inside the innermost loop is not contigious, and the
target hardware doesn't support efficient gather operations,
\loopnest{} will fallback to scalar operations.

\subsubsection*{No--Spilling of the Resulting Tensor}

The concept of tiling (or "blocking") for the multiple levels of a
cache hierarchy and the register file is a well known optimization
technique~\cite{goto2008anatomy, gunnels2001family, lam1991cache,
  smith2014anatomy}; referred to as \emph{Cache Blocking Techniques}
in the Intel's optimization manual~\cite{intel2014intel}. \loopnest{}
exposes this common HPC optimization, where the subset of the output
tensor is kept in register file~\cite{onednn, heinecke2015libxsmm,
  jia2018optimizing, zlateski2018deeper, zlateski2019anatomy,
  heinecke2016libxsmm, heinecke2016high}.  \loopnest{} never produces
code that spills the content of the register file to the stack, an
approach that is commonly used in traditional compilers, such as LLVM
or GCC.  Spilling the content of the register file to stack puts
pressure on the hardware's load/store units, preventing full hardware
utilization.

\loopnest{} does not decide on blocking or tiling sizes, nor on the
size used for the data kept in registers (register
blocking). \loopnest{} instead identifies the outermost user provided
loop for which the all compute can be performed with a subset of the
output tensor kept in registers.  Thus, it's up to the user to provide
a well chosen loop order and sizes -- one where the values kept in the
register file can be reused many times.  This approach gives the user
a greater control, with more predictable performance as compared to
the case when spilling is allowed.

\subsection{Single Operand LoopNest}

\loopnest{} also provides functionality for generating efficient code
for a simplified tensor contraction, where there are no reduction
dimensions, and only one input is provided.  This functionality is
typically used for \textit{reshaping} a tensor (such as numpy's
\textbf{reshape} function), but can also be used for extracting a
subset of a tensor into a smaller tensor, or broadcasting elements
along a tensor dimension.  This functionality is required by
\loopstack{} in order to allow the user's schedules that prefer to
reorganize the memory for faster access~\cite{goto2008anatomy,
  gunnels2001family, lam1991cache, smith2014anatomy}

\subsection{Generalizing to computations with multiple loop nests}

Some computations or their schedules can result in a sequence of
multiple nested loops that may share a set of outer loops, effectively
forming a tree of loops. To generalize our approach to these
workloads, we developed a loop tree interface. The loop tree interface
provides a simple API to build up a tree, where inner nodes correspond
to for-loops, and leaves correspond to an innermost computation over
tensors or a transposition of tensors. \loopnest{} then compiles all
independent nests and executes the tree. The final result is a
function that can be called with the appropriate input, intermediate,
and output tensors to realize the computation defined by the tree.

\section{IR Manipulation and Optimization}
\label{optimization}

\subsection{Interaction Model}

A core challenge in high-performance tensor operations is identifying
a good execution order -- the schedule. As noted previously, there can
be many valid schedules for a given computation: loops can be split
and reordered, dimensions can be split and swapped, intermediate
tensors can be packed, and so on. As a result, any scheduling tool is
faced with the challenge of an exponentially large search space making
simple enumeration infeasible. Recent work~\cite{zheng2020ansor,
  adams2019learning} has explored using a combination of machine
learning and structured search strategies to automatically explore the
search space.

While automation presents a promising direction, expert intuitions
(and their use in a guided search) remain an important factor in
developing high performance kernels today. As a result, a key goal of
\loopstack{} is to facilitate exploration of schedules. Effective
exploration depends on real-time feedback, which an expert can use to
validate their intuitions and incrementally explore the space of
possible schedules. To deliver real-time,
instantaneous~\cite{nielsen1994usability}, feedback, \looptool{}
exploits \loopnest{}'s fast code generation to compile and benchmark
schedules with low-latency (on the order of 10-100ms for programs not
bounded by the execution of the underlying generated code).

\subsection{Manual Tuning}

\begin{figure}
    \flushleft
    \begin{subfigure}{1.0\columnwidth}
    \includegraphics[width=\columnwidth]{\manualstepsfigpath{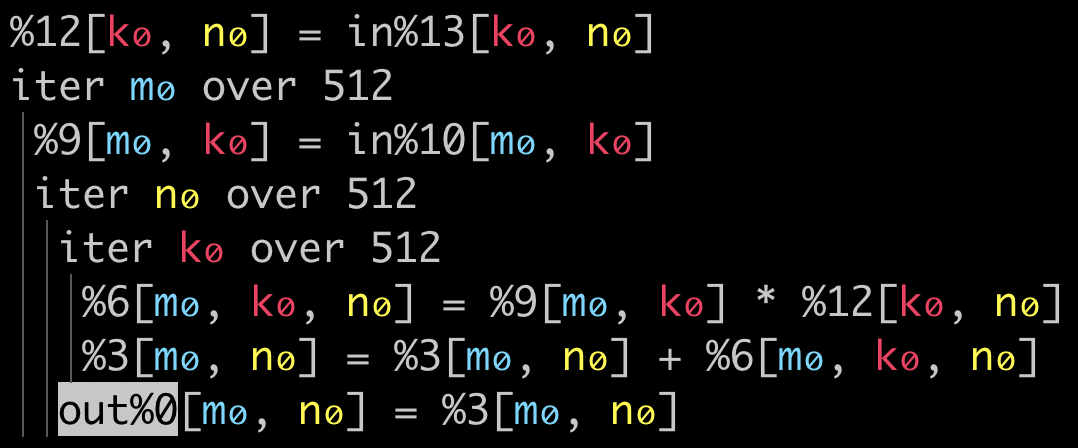}}
    \end{subfigure}
    \begin{subfigure}{\columnwidth}
    \includegraphics[width=\columnwidth]{\manualstepsfigpath{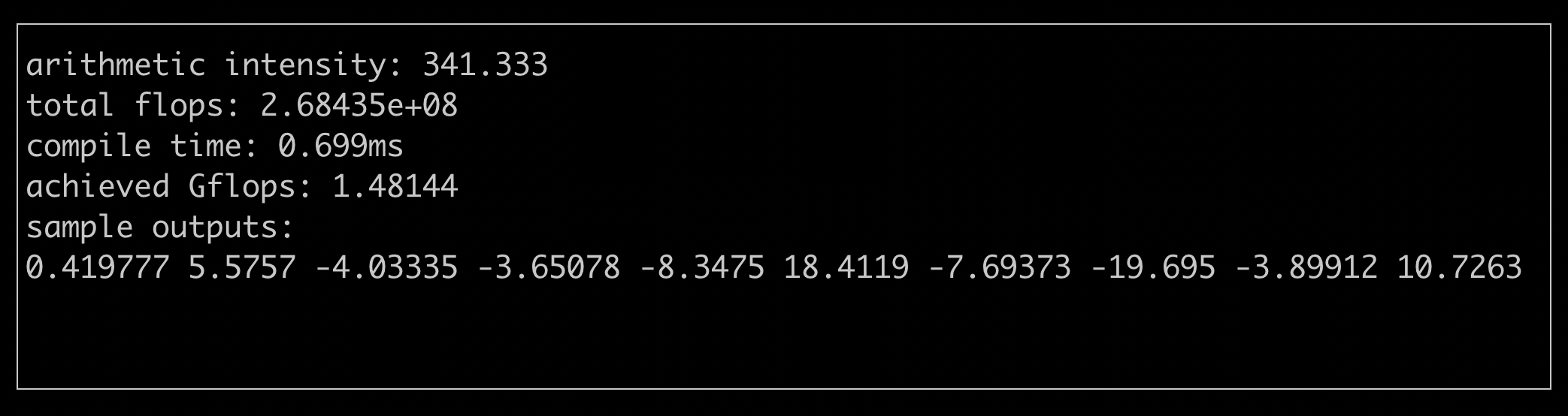}}
    \end{subfigure}
    \caption{Initial UI for a 512x512x512 matrix multiplication and a
      real-time measured performance window.}
    \label{fig:manual-demo-bench-0}
\end{figure}

\begin{figure}
    \flushleft
    \begin{subfigure}{\columnwidth}
    \includegraphics[width=\columnwidth]{\manualstepsfigpath{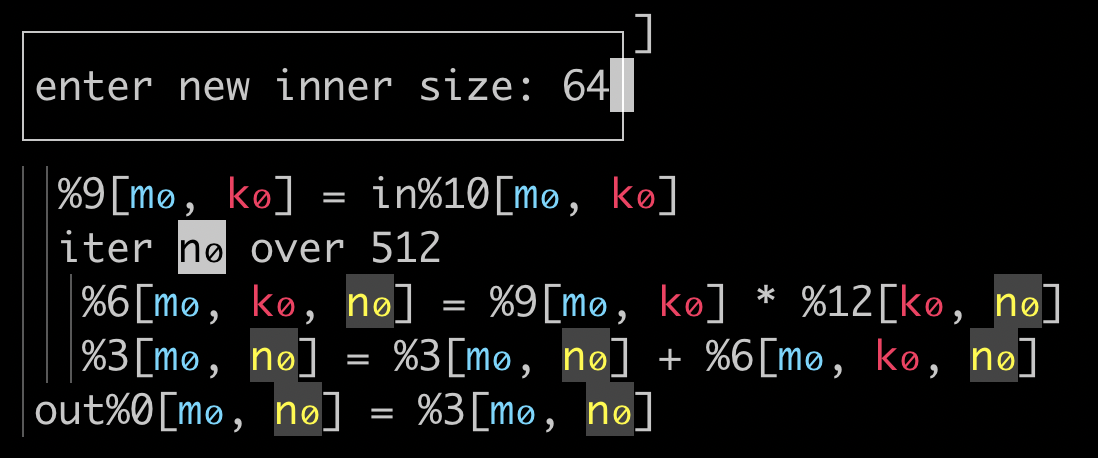}}
    \end{subfigure}
    \begin{subfigure}{\columnwidth}
    \includegraphics[width=\columnwidth]{\manualstepsfigpath{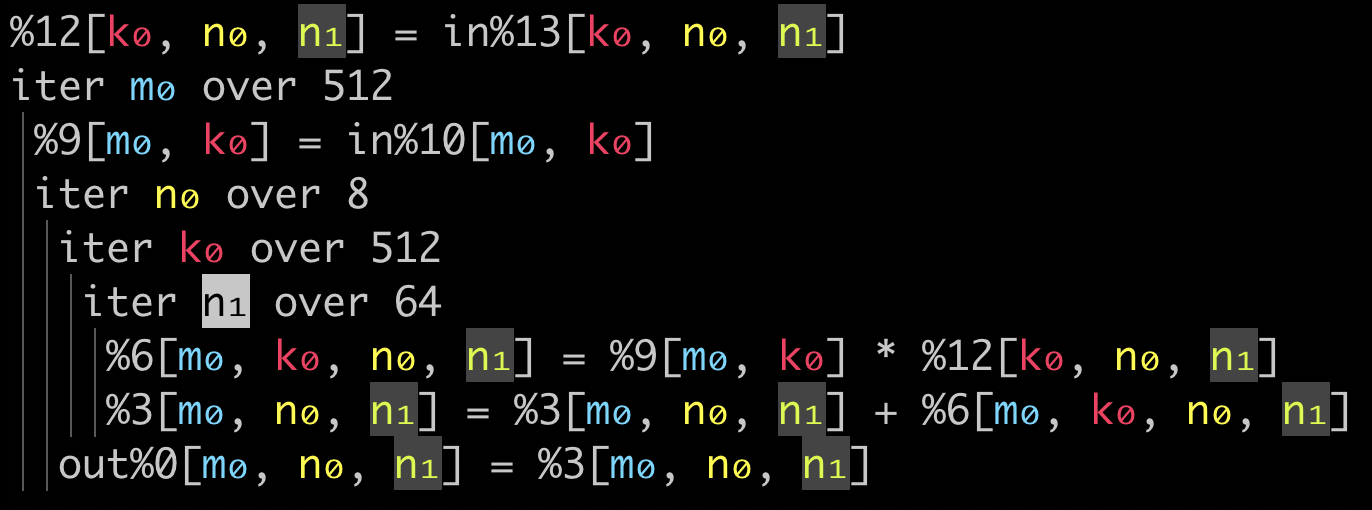}}
    \end{subfigure}
    \begin{subfigure}{\columnwidth}
    \includegraphics[width=\columnwidth]{\manualstepsfigpath{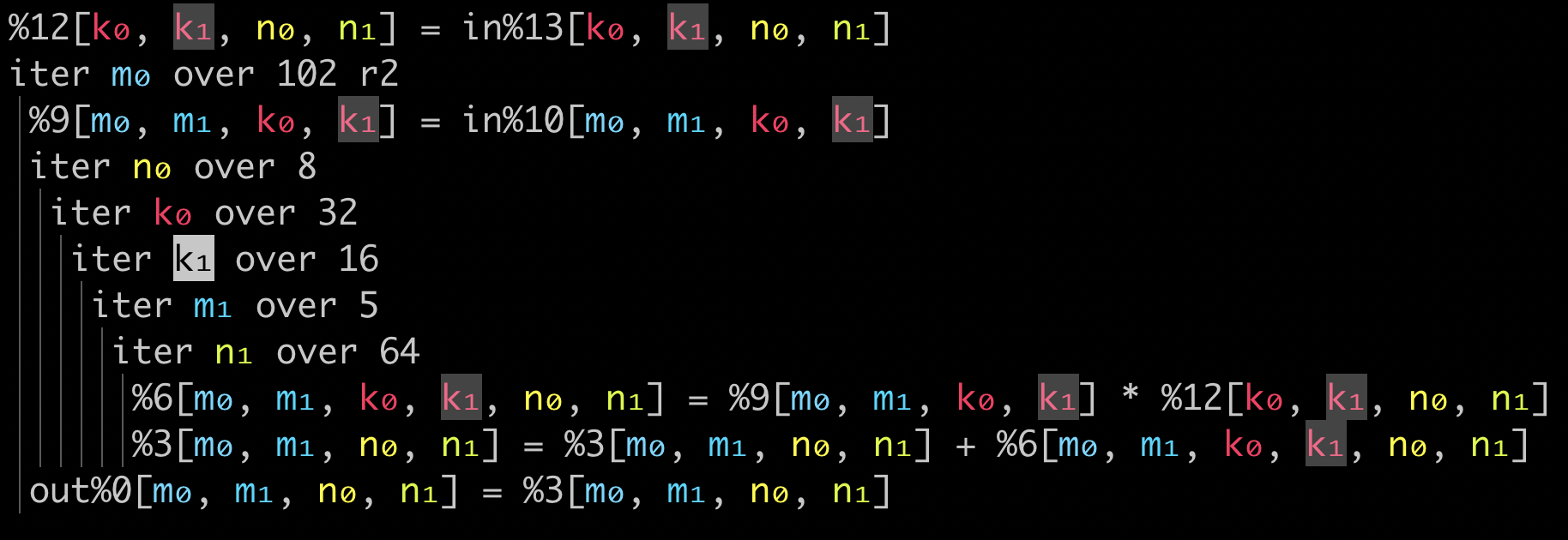}}
    \end{subfigure}
    \begin{subfigure}{\columnwidth}
    \includegraphics[width=\columnwidth]{\manualstepsfigpath{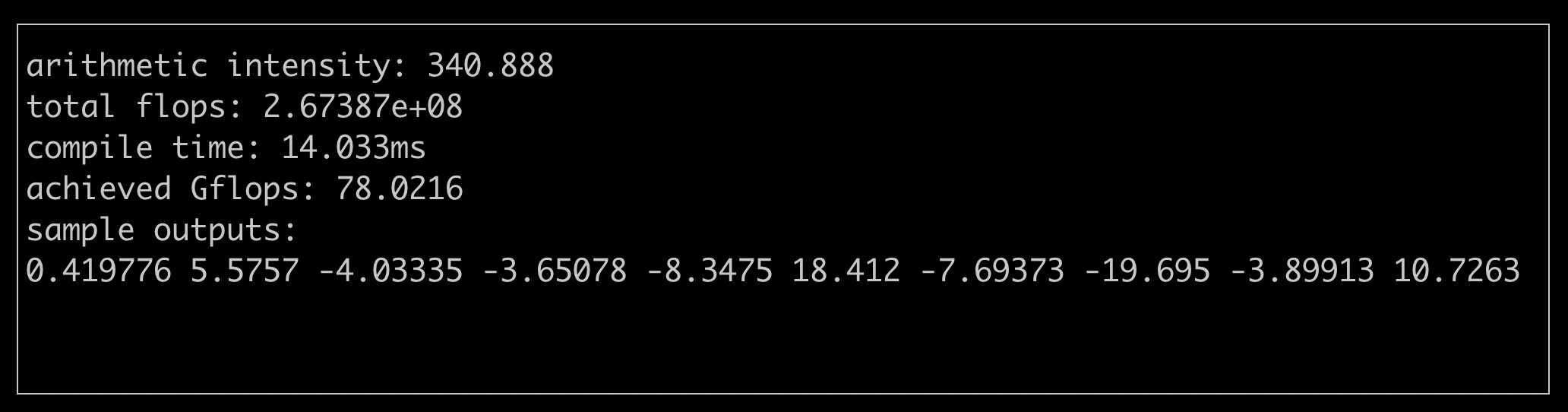}}
    \end{subfigure}
    \caption{Splitting loops vastly improves performance.
      \loopstack{} gracefully handles tail logic (seen for variable
      \texttt{m}), enabling efficient 5x64 register blocking. The
      \texttt{k} loop is split to induce unrolling.}
    \label{fig:manual-demo-bench-1}
\end{figure}

\begin{figure}
    \flushleft
    \begin{subfigure}{\columnwidth}
    \includegraphics[width=\columnwidth]{\manualstepsfigpath{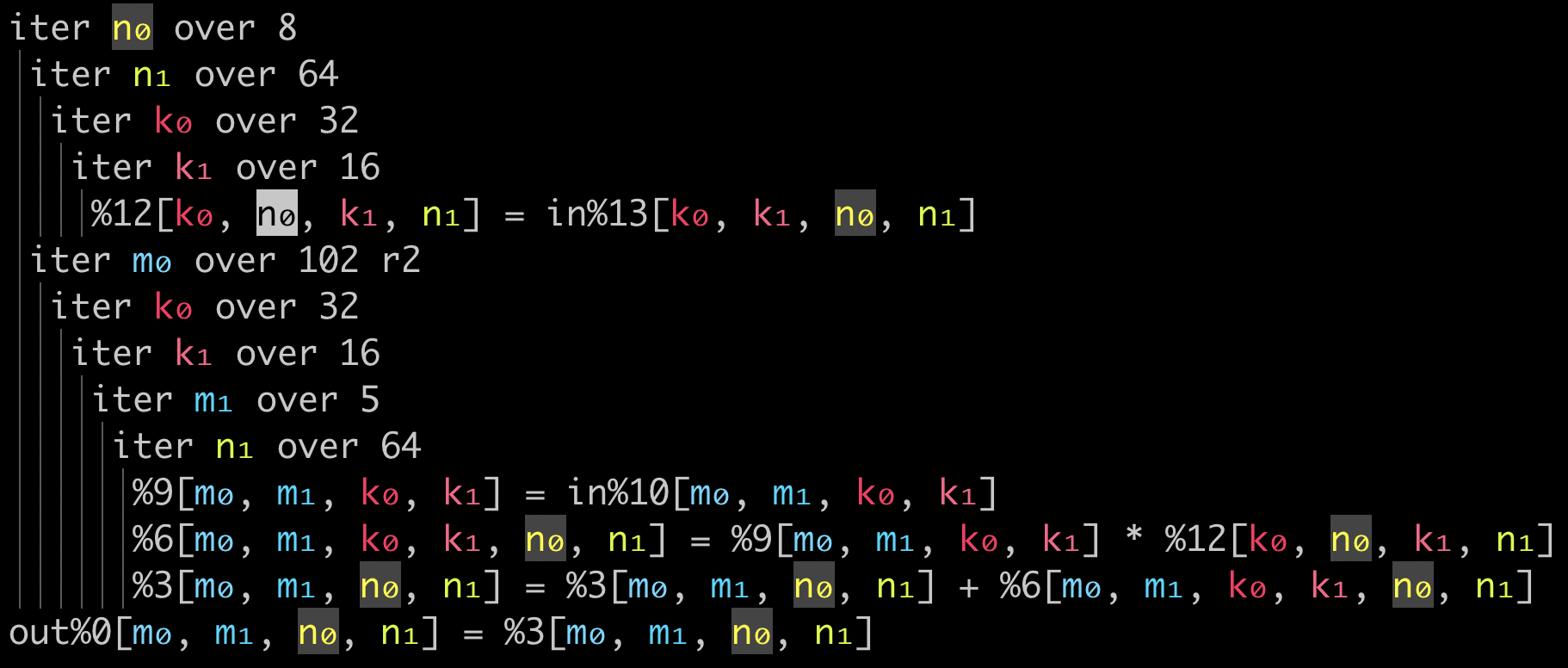}}
    \end{subfigure}
        \begin{subfigure}{\columnwidth}
    \includegraphics[width=\columnwidth]{\manualstepsfigpath{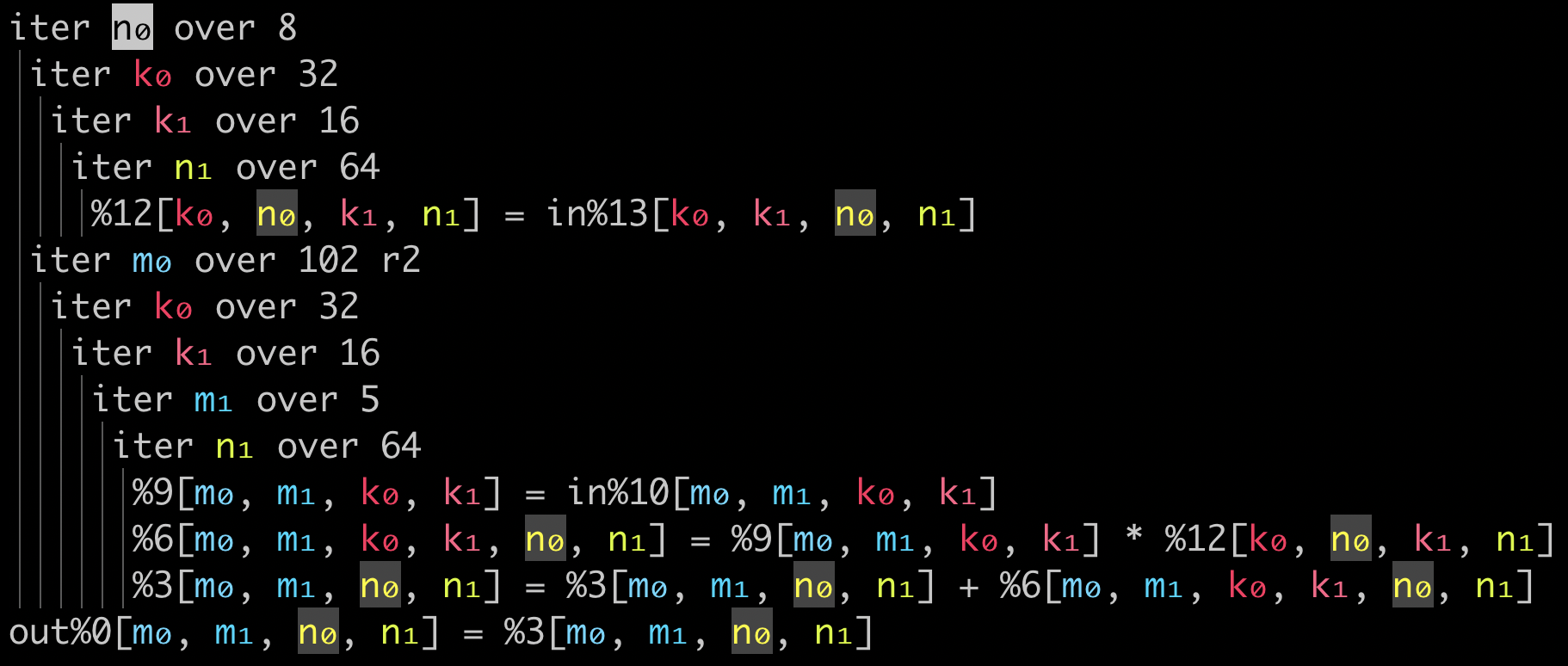}}
    \end{subfigure}
    \begin{subfigure}{\columnwidth}
    \includegraphics[width=\columnwidth]{\manualstepsfigpath{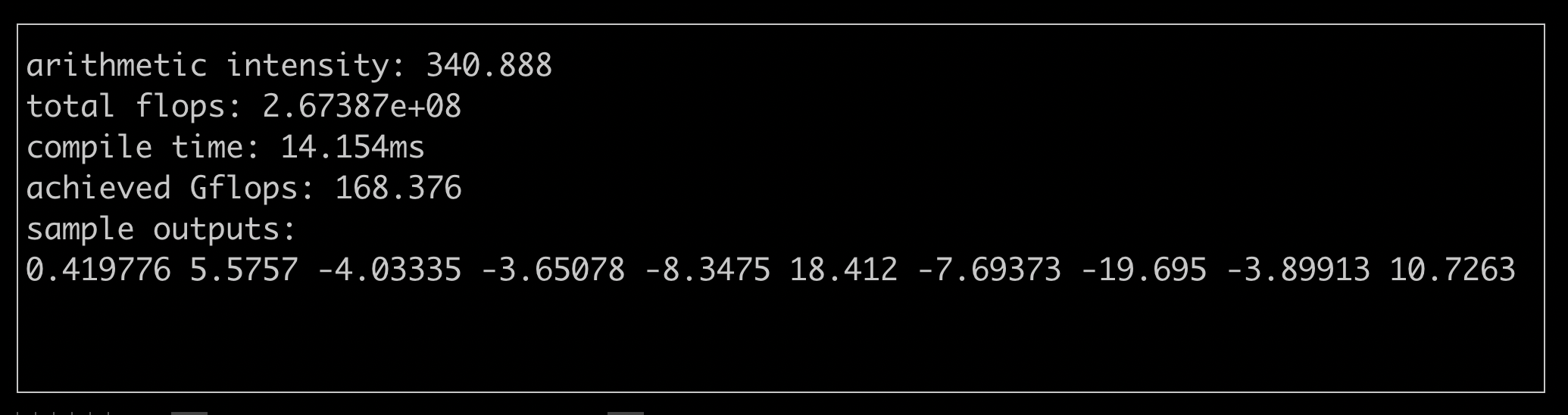}}
    \end{subfigure}
    \caption{Reordering the virtual memory at \texttt{\%12} (packing)
      induces a new loop that also needs to be optimized.  Because the
      loop over \texttt{n\_0} is shared, only 32KB of scratch space is
      used for \texttt{\%12}.}
    \label{fig:manual-demo-bench-2}
 
\end{figure}

To promote human driven exploration with \looptool{}, we developed a
terminal based user interface
(TUI). Fig.~\ref{fig:manual-demo-bench-0} presents a matrix
multiplication example using \looptool{}'s visual TUI. The schedule
for the computation is reflected in the TUI and provides controls to
interactively manipulate the schedule, including splitting and
reordering loops and memory.  \looptool{} can automatically benchmark
the schedule presented and display statistics useful for
optimization. As shown in Fig.~\ref{fig:manual-demo-bench-0},
\looptool{} reports the total required number of operations (FLOPs),
arithmetic intensity (The ratio of FLOPs performed and memory
transfers~\cite{williams2008roofline, williams2008roofline,
  ilic2014cache}), and the achieved performances in billions of
floating point operations per second (GFLOPS). This information allows
experts to iterate on kernel schedules quickly, and non-experts to
ramp up on schedule and hardware performance characteristics.

The speed with which a developer can experiment with different
schedules allows exploring solutions optimally tuned for the
computation and hardware at hand. For example, \looptool{} can quickly
show the impact of different layouts for inputs or output, or the use
of intermediate memory (shared across sequentially composed kernels)
sized for the system's cache.

An example of manual optimizations on an AVX512 capable CPU with peak
performances of around 190 GFLOPS can be seen in
figures~\ref{fig:manual-demo-bench-0}, ~\ref{fig:manual-demo-bench-1},
and ~\ref{fig:manual-demo-bench-2}. The initial, trivial schedule, as
well as the poor performances (less than 1\% utilization) achieved,
are shown on Fig.~\ref{fig:manual-demo-bench-0}.  The
Fig.~\ref{fig:manual-demo-bench-1} shows the process of splitting and
reordering loops, which resulted in around 40\% utilization.  Finally,
the Fig.~\ref{fig:manual-demo-bench-2} shows the process of
introducing memory packing, resulting in a final schedule that results
in utilization of 88.5\% on average (over many invocations of the
compiled schedule).

\subsection{Scripted Tuning}
\label{tuning-section}

As we have discussed above, an efficient tensor contraction schedule
will use cache/register blocking (also referred to as tiling), a
technique that improves performance by increasing data locality that
has been studied for forever~\cite{stone1970logic, wolfe1989more,
  brown1975numerical, wolf1991data, irigoin1988supernode,
  bouwmeester2012tiled, kurt2020efficient, hong2019adaptive,
  wiki:lnopt, metcalf1980fortran}.  An expert, familiar with such
techniques, as well as the design and limitations of modern hardware
can easily script a reasonable sweep (on the order of hundreds to
thousands of schedules) of parameterizations of these common
techniques.

To demonstrate this, we wrote a simple three step automatic tuning
script.  Clocking in under 1000 lines of code, this script sweeps
through possible tile sizes that are expected to perform well on
modern hardware (as presented in~\cite{goto2008anatomy,
  smith2014anatomy}) and runs on the order of seconds to minutes on a
single core for most individual neural network operations or blocks.
The script works as follows.

First we define a set of tile sizes (based on modern hardware's cache
sizes) that will be used in conjunction with a set of register
blocking sizes (based on current hardware's register file size).  This
approach mimics the last two levels of what both
Goto~\cite{goto2008anatomy} and Smith~\cite{smith2014anatomy} refer to
"Anatomy of High--Performance Matrix multiplication".  Given the size
of the register file of a given hardware, we sweep over different
\textit{register blocking} sizes that fit in the register file.  For
each such \textit{register blocking} size, we also sweep over blocking
sizes (as well as order of compute~\cite{stone1970logic,
  wolfe1989more, brown1975numerical, wolf1991data,
  irigoin1988supernode, bouwmeester2012tiled, kurt2020efficient,
  hong2019adaptive, wiki:lnopt, metcalf1980fortran}).  We then
benchmark every combination of tile size and register block, saving
the best configuration for the next step.

In the second step we partially sweep the set of every permutation of
the loop order.  At this point, with both tiling and register
blocking, the loop nest tends to be on the order of 10 or so deep,
which makes a full sweep infeasible (given our time constraint of
minutes or less).  We approximate tuning for every permutation by
instead permuting windows of the loop nest.  Starting with the
innermost 5 loops, we try all combinations keeping the outermost loops
fixed.  We then continue this process upward, saving the best nesting
order each time.

Finally, we attempt to inject memory packing to increase cache use.
This involves a mix of permuting virtual memory layouts to align with
later reads and writes and staging loops to induce memory usage.  This
helps with processor architectures that benefit from memory speed and
problem sizes that are compute bound with large inputs.

It is important to note that the above optimizations vastly reduce the
search space of schedules by relying on linearizing assumptions.  We
feel the time saved during the search is a good compromise that
enables realistic use of tuning in research or production settings.

\section{Evaluation}

We performed a series of experiments in order to answer the following
questions:

\begin{itemize}
\item How does the compile time of \loopnest{} compare to the one of
  traditional compilers, such as LLVM?
\item How does the generated code perform as compared to one obtained
  with traditional compilers?
\item How does code generated by \looptool{}{} compare against state
  of the art ML primitives, as well as end--to--end models?
\item How easy is it to extend \looptool{} to support new sets of
  instructions?
\end{itemize}

\subsection{Setup}

Experiments are performed on total of 6 different CPU cores and 4
different vector extensions: Intel Xeon Platinum 8124M with the avx512
vector extension; AMD Epyc 7R32, with the avx2 vector extension;
Cortex A57, Cortex A73, NVIDIA Denver 2 and Apple M1 CPUs -- all with
ARM's neon vector extension; and finally, Apple M1 with the neon\_fp16
vector extension.  We perform all experiments single--threaded; using
a single core on the target hardware, in order to analyze the
contributions of our work. 
Using multi--threaded workloads would also depend on the underlying
threading implementations, and can skew our desired measurements.

\subsection{Operator Benchmarking vs Traditional Compilers}

A key contribution of \loopnest{}, the backend of \loopstack{}, is the
reduced compile times required for generating high performance
code. To validate this contribution, we compare \loopstack{} to LLVM,
which is a popular backend choice for existing tensor operation
compilers such as Halide~\cite{ragan2013halide, steiner2021value} and
TVM~\cite{tvm, autotvm}.

Figure~\ref{table:operator-benchmarks} presents 12 operator we have
evaluated, consisting of matrix multiplications (fully connected
layers), convolutions, and depth-wise convolutions over varying input
sizes. To provide a fair comparison of LLVM and \loopstack{}, we
evaluate both systems on the same set of schedules and perform
pair-wise comparisons. For LLVM code generation, we use Halide to emit
schedules identical to the ones used with \loopstack{}. We ensure that
LLVM's runtime assertions, as well as bound checks, are turned off. As
LLVM is a general, \textit{catch--all} compiler, performing
apples--to--apples comparison is not an easy task; for that reason, we
instruct LLVM to use the same optimizations as the ones used in
practice -- such as the ones used in Halide~\cite{gareev2018high,
  steiner2021value, adams2019learning} and TVM~\cite{tvm, autotvm}.
We focus our evaluation on two key dimensions: compile time and
execution time.

We generate a set of schedules with the rule-based heuristic scheduler
referred to in Section~\ref{tuning-section}, which performs a sweep of
execution schedule optimizations such as tiling. We attempt to collect
schedules seen throughout this tuning process, but only collect
schedules that result in a single loop nest (i.e. there are no shared
outer loops across different nested inner loops). After we obtain a
full set of schedules for a benchmark, we evaluate both \loopstack{}
and LLVM on them.

\begin{table}[]
\resizebox{\columnwidth}{!}{
\begin{tabular}{@{}lll@{}}
\toprule
Operator               & Sizes           & Benchmark Name \\ \midrule
GEMM                   & 64 x 64 x 64 & MM-64         \\
GEMM                   & 128 x 128 x 128 & MM-128         \\
GEMM                   & 256 x 256 x 256 & MM-256         \\
GEMM                   & 512 x 512 x 512 & MM-512         \\
Convolution            & (channels:64 to 128), size:56x56, filter:3x3, stride:1 & CONV-1          \\
Convolution            & (channels:128 to 256), size:28x28, filter:3x3, stride:1 & CONV-2          \\
Convolution            & (channels:256 to 512), size:14x14, filter:3x3, stride:1 & CONV-3          \\
Convolution            & (channels:512 to 512), size:7x7, filter:3x3, stride:1 & CONV-4          \\
Depth-wise Convolution & (channels:16), size:112x112, filter:3x3, stride:2 & DWCONV-1          \\
Depth-wise Convolution & (channels:72), size:56x56, filter:3x3, stride:2 & DWCONV-2          \\
Depth-wise Convolution & (channels:88), size:28x28, filter:3x3, stride:1 & DWCONV-3          \\
Depth-wise Convolution & (channels:240), size:14x14, filter:5x5, stride:1 & DWCONV-4          \\
\bottomrule
\end{tabular}
}
\caption{Summary of operator benchmarks for \loopnest{}/LLVM comparison}
\label{table:operator-benchmarks}
\end{table}

Figure~\ref{table:operator-benchmarks-compile-tables} presents a
summary of compile times across benchmarks.  The ``LLVM'' and
``\loopnest{} (LN)'' columns present the average number of
milliseconds to generate machine code for schedules for the
corresponding benchmark. The ``ratio'' columns presents the average
ratio of LLVM compile time relative to LoopNest compile time. Our
experiments show that \loopnest{}'s code generation is faster than
LLVM's compilation for all schedules, all workloads and on all target
hardware.  In most cases \loopnest{} can generate code orders of
magnitude faster.  The significantly shorter compilation times enable
use of \loopnest{} in automated schedule searches.

\begin{table*}[t]
\resizebox{\textwidth}{!}{
    \centering
    \setlength\tabcolsep{2.5pt}
    \begin{tabular}{r !{\vrule width0.8pt} rrr|rrr !{\vrule width0.8pt} rrr|rrr|rrr|rrr  }
      & \multicolumn{6}{c !{\vrule width0.8pt} }{\textbf{x86 based CPUs}} & \multicolumn{12}{c}{\textbf{Aarch64 (ARM) based CPUs (NEON)}} \\
      & \multicolumn{3}{c|}{AMD (AVX2)} & \multicolumn{3}{c!{\vrule width0.8pt}}{Intel (AVX512)}
      & \multicolumn{3}{c|}{Cortex A57} & \multicolumn{3}{c|}{NVIDIA Denver2}
      & \multicolumn{3}{c|}{Cortex A73} & \multicolumn{3}{c}{Apple M1} \\
      & LLVM & LN & Ratio & LLVM & LN & Ratio & LLVM & LN & Ratio 
      & LLVM & LN & Ratio & LLVM & LN & Ratio & LLVM & LN & Ratio \\
      \hline
CONV-1 & 820.78 & {\bf 2.5153} & {\bf 326.31} & 9881.9 & {\bf 6.2062} & {\bf 1592.3} & 2948.4 & {\bf 9.2109} & {\bf 320.1} & 3972.4 & {\bf 30.28} & {\bf 131.19} & 4914.6 & {\bf 22.077} & {\bf 222.61} & 555.24 & {\bf 24.763} & {\bf 22.422}\\ 
CONV-2 & 1590.1 & {\bf 11.717} & {\bf 135.7} & 9531.3 & {\bf 5.8035} & {\bf 1642.3} & 2481.1 & {\bf 8.9475} & {\bf 277.29} & 4798.5 & {\bf 21.768} & {\bf 220.44} & 4310.1 & {\bf 31.053} & {\bf 138.8} & 531.12 & {\bf 15.261} & {\bf 34.802}\\ 
CONV-3 & 762.52 & {\bf 5.9325} & {\bf 128.53} & 11061 & {\bf 11.411} & {\bf 969.29} & 3113.2 & {\bf 17.749} & {\bf 175.4} & 4184.3 & {\bf 24.184} & {\bf 173.02} & 3919.1 & {\bf 25.51} & {\bf 153.63} & 444.29 & {\bf 20.435} & {\bf 21.742}\\ 
CONV-4 & 885.74 & {\bf 41.163} & {\bf 21.518} & 10706 & {\bf 14.264} & {\bf 750.56} & 4084.3 & {\bf 66.102} & {\bf 61.788} & 5169.9 & {\bf 85.97} & {\bf 60.136} & 6408.8 & {\bf 99.952} & {\bf 64.119} & 827.22 & {\bf 35.605} & {\bf 23.233}\\ 
DWCONV-1 & 838.46 & {\bf 0.29416} & {\bf 2850.3} & 10551 & {\bf 0.28027} & {\bf 37647} & 2775.9 & {\bf 3.8011} & {\bf 730.28} & 4108.2 & {\bf 5.0464} & {\bf 814.09} & 4844.8 & {\bf 2.8047} & {\bf 1727.4} & 571.46 & {\bf 1.209} & {\bf 472.67}\\ 
DWCONV-2 & 1033.8 & {\bf 0.47162} & {\bf 2192} & 11812 & {\bf 0.67874} & {\bf 17402} & 2795.3 & {\bf 2.5241} & {\bf 1107.4} & 3814.5 & {\bf 5.2036} & {\bf 733.06} & 4160.9 & {\bf 1.8825} & {\bf 2210.3} & 470.53 & {\bf 0.5241} & {\bf 897.79}\\ 
DWCONV-3 & 969.88 & {\bf 0.30031} & {\bf 3229.6} & 13759 & {\bf 1.4601} & {\bf 9423.3} & 2726.6 & {\bf 2.158} & {\bf 1263.5} & 3666.1 & {\bf 4.2326} & {\bf 866.15} & 3320 & {\bf 3.8216} & {\bf 868.74} & 506.63 & {\bf 0.66237} & {\bf 764.87}\\ 
DWCONV-4 & 1000.7 & {\bf 0.33429} & {\bf 2993.4} & 10704 & {\bf 0.81343} & {\bf 13159} & 3474 & {\bf 5.5112} & {\bf 630.34} & 4577.5 & {\bf 9.4771} & {\bf 483.01} & 4197.4 & {\bf 2.8767} & {\bf 1459.1} & 449.89 & {\bf 1.6} & {\bf 281.18}\\ 
MM-64 & 697.37 & {\bf 0.38452} & {\bf 1813.6} & 9099.5 & {\bf 0.85309} & {\bf 10667} & 3432.4 & {\bf 7.3588} & {\bf 466.43} & 4779.2 & {\bf 13.103} & {\bf 364.75} & 4417.6 & {\bf 9.1708} & {\bf 481.7} & 397.59 & {\bf 1.3267} & {\bf 299.68}\\ 
MM-128 & 925.47 & {\bf 1.578} & {\bf 586.47} & 9044.8 & {\bf 0.79484} & {\bf 11379} & 4991.9 & {\bf 11.153} & {\bf 447.57} & 5754.2 & {\bf 16.025} & {\bf 359.07} & 6681.1 & {\bf 13.956} & {\bf 478.72} & 387.82 & {\bf 1.1188} & {\bf 346.63}\\ 
MM-256 & 1118.5 & {\bf 2.6925} & {\bf 415.41} & 18119 & {\bf 3.0202} & {\bf 5999.4} & 5003.1 & {\bf 18.511} & {\bf 270.28} & 5770.9 & {\bf 26.485} & {\bf 217.9} & 6800.6 & {\bf 27.446} & {\bf 247.78} & 390.29 & {\bf 5.1647} & {\bf 75.57}\\ 
MM-512 & 1262.3 & {\bf 4.3405} & {\bf 290.83} & 12336 & {\bf 4.4847} & {\bf 2750.8} & 4814.2 & {\bf 7.4852} & {\bf 643.17} & 5698.9 & {\bf 12.25} & {\bf 465.22} & 6471.7 & {\bf 14.545} & {\bf 444.94} & 1084.4 & {\bf 9.5121} & {\bf 114} \\
      \hline
    \end{tabular}
    }
\caption{
Compile times, in milliseconds, for {\ttfamily LLVM} and
\loopnest{}. \loopnest{} performs the compilation orders of magnitude faster}
\label{table:operator-benchmarks-compile-tables}
\end{table*}

In Figure~\ref{table:operator-benchmarks-compile-tables}, we show the
average run--times of the top 5 fastest schedules, for each compiler.
While LLVM does outperform on some very inefficient schedules (not
presented), we decide not to use slower schedules in our analysis, as
such schedules will not be used in practice.  Our results suggest that
\loopnest{} is generating code that has at least comparable
performances, or exceeds the performances of the one generated by LLVM
-- while taking just a fraction of time.  Additionally, as a
dependency, \loopstack{}'s carries only a binary of $250Kb$ in size,
as compared to LLVM (which defaults to more than $350Mb$); this makes
\loopnest{} an obvious choice for at least mobile and embedded
systems.

\begin{table*}[t]
\resizebox{\textwidth}{!}{
    \centering
    \setlength\tabcolsep{2.5pt}
    \begin{tabular}{r !{\vrule width0.8pt} rrr|rrr !{\vrule width0.8pt} rrr|rrr|rrr|rrr  }
      & \multicolumn{6}{c !{\vrule width0.8pt} }{\textbf{x86 based CPUs}} & \multicolumn{12}{c}{\textbf{Aarch64 (ARM) based CPUs (NEON)}} \\
      & \multicolumn{3}{c|}{AMD (AVX2)} & \multicolumn{3}{c!{\vrule width0.8pt}}{Intel (AVX512)}
      & \multicolumn{3}{c|}{Cortex A57} & \multicolumn{3}{c|}{NVIDIA Denver2}
      & \multicolumn{3}{c|}{Cortex A73} & \multicolumn{3}{c}{Apple M1} \\
      & LLVM & LN & Ratio & LLVM & LN & Ratio & LLVM & LN & Ratio 
      & LLVM & LN & Ratio & LLVM & LN & Ratio & LLVM & LN & Ratio \\
      \hline
CONV-1 & 86.767 & {\bf 87.638} & {\bf 1.01} & 72.943 & {\bf 162.48} & {\bf 2.2274} & 11.021 & {\bf 11.126} & {\bf 1.0095} & 10.968 & {\bf 14.782} & {\bf 1.3477} & {\bf 9.8466} & 9.3994 & 0.95458 & {\bf 98.121} & 97.51 & 0.99377\\ 
CONV-2 & 45.765 & {\bf 71.482} & {\bf 1.5619} & 13.813 & {\bf 156.78} & {\bf 11.35} & {\bf 11.276} & 11.179 & 0.99134 & 13.407 & {\bf 14.511} & {\bf 1.0824} & {\bf 9.8105} & 9.2468 & 0.94254 & {\bf 95.743} & 92.027 & 0.96119\\ 
CONV-3 & 9.4946 & {\bf 72.433} & {\bf 7.6289} & 96.448 & {\bf 184.4} & {\bf 1.9119} & 8.1811 & {\bf 10.183} & {\bf 1.2447} & 6.4069 & {\bf 13.27} & {\bf 2.0712} & 6.4671 & {\bf 7.9984} & {\bf 1.2368} & 53.681 & {\bf 83.252} & {\bf 1.5509}\\ 
CONV-4 & 3.307 & {\bf 90.722} & {\bf 27.434} & 123.73 & {\bf 181.72} & {\bf 1.4687} & 5.3246 & {\bf 9.5687} & {\bf 1.7971} & 4.3611 & {\bf 12.532} & {\bf 2.8736} & 2.9514 & {\bf 7.686} & {\bf 2.6042} & 46.806 & {\bf 77.952} & {\bf 1.6654}\\ 
DWCONV-1 & 48.695 & {\bf 62.541} & {\bf 1.2844} & 48 & {\bf 57.592} & {\bf 1.1998} & {\bf 6.3521} & 6.2343 & 0.98146 & 10.243 & {\bf 11.858} & {\bf 1.1577} & 3.9988 & {\bf 4.5019} & {\bf 1.1258} & {\bf 40.133} & 39.01 & 0.97201\\ 
DWCONV-2 & 39.203 & {\bf 53.465} & {\bf 1.3638} & 28.302 & {\bf 37.671} & {\bf 1.331} & 5.1196 & {\bf 5.7031} & {\bf 1.114} & 7.1075 & {\bf 7.9688} & {\bf 1.1212} & 2.5178 & {\bf 2.7729} & {\bf 1.1013} & 42.865 & {\bf 43.438} & {\bf 1.0134}\\ 
DWCONV-3 & 62.873 & {\bf 84.848} & {\bf 1.3495} & 57.544 & {\bf 88.094} & {\bf 1.5309} & 6.9338 & {\bf 7.7209} & {\bf 1.1135} & 10.65 & {\bf 12.551} & {\bf 1.1785} & 5.7644 & {\bf 6.4571} & {\bf 1.1202} & 70.796 & {\bf 76.261} & {\bf 1.0772}\\ 
DWCONV-4 & 77.071 & {\bf 84.21} & {\bf 1.0926} & 125.04 & {\bf 159.38} & {\bf 1.2747} & 9.6875 & {\bf 10.174} & {\bf 1.0502} & 12.665 & {\bf 14.717} & {\bf 1.1621} & 7.6075 & {\bf 8.0169} & {\bf 1.0538} & {\bf 81.588} & 80.901 & 0.99157\\ 
MM-64 & 85.808 & {\bf 102.2} & {\bf 1.191} & 144.67 & {\bf 187.65} & {\bf 1.2971} & 12.751 & {\bf 13.441} & {\bf 1.0541} & 14.039 & {\bf 15.754} & {\bf 1.1221} & 9.523 & {\bf 10.166} & {\bf 1.0675} & 91.18 & {\bf 199.81} & {\bf 2.1913}\\ 
MM-128 & 92.692 & {\bf 102.5} & {\bf 1.1058} & 168.84 & {\bf 185.19} & {\bf 1.0969} & 12.417 & {\bf 12.496} & {\bf 1.0064} & 14.253 & {\bf 15.291} & {\bf 1.0728} & 9.2462 & {\bf 9.5062} & {\bf 1.0281} & 91.763 & {\bf 98.4} & {\bf 1.0723}\\ 
MM-256 & 92.862 & {\bf 100.21} & {\bf 1.0791} & 170.18 & {\bf 182.46} & {\bf 1.0722} & {\bf 11.428} & 11.401 & 0.99761 & 14.241 & {\bf 14.645} & {\bf 1.0284} & {\bf 8.6863} & 8.6103 & 0.99125 & 95.597 & {\bf 99.902} & {\bf 1.045}\\ 
MM-512 & 90.189 & {\bf 98.199} & {\bf 1.0888} & {\bf 160.42} & 159.59 & 0.9948 & 6.9971 & {\bf 8.3267} & {\bf 1.19} & {\bf 14.395} & 13.894 & 0.9652 & 6.3364 & {\bf 6.8905} & {\bf 1.0874} & 89.618 & {\bf 97.65} & {\bf 1.0896}\\
      \hline
    \end{tabular}
    }
\caption{ Meeasured execution performances, in GFLOPS, for code
  generatied by {\ttfamily LLVM} and \loopnest{}. \loopnest{} achieves
  comparable (within measurement error) or superior performances,
  while taking a fraction of time to generate the code.}
\label{table:operator-benchmarks-runtime-tables}
\end{table*}

\begin{figure}
\includegraphics[width=0.91\columnwidth]{\versusLLVMfigpath{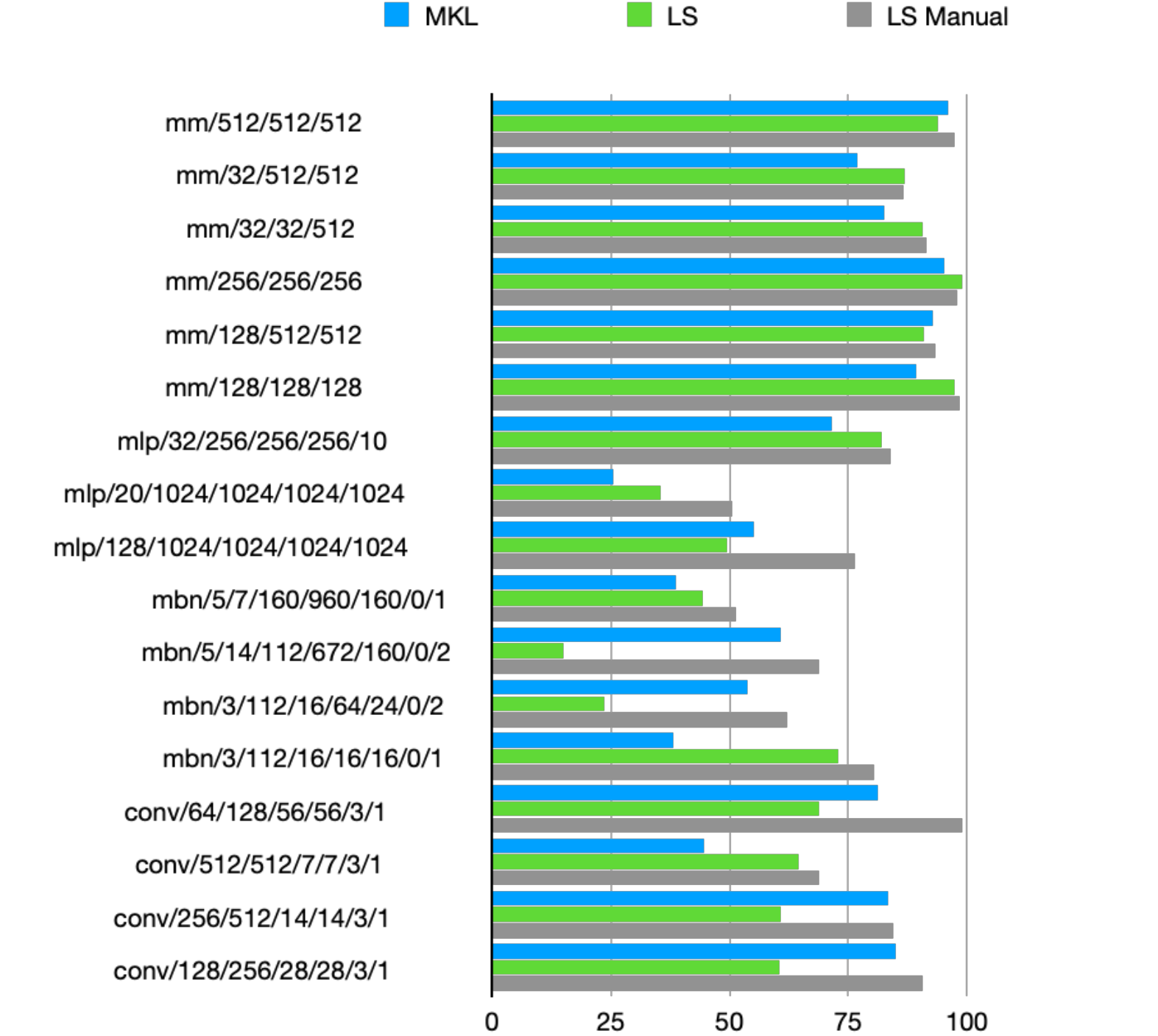}}
\caption{Comparison of MKL-DNN, Scripted, and Manual tuning using \looptool{} on an AMD EPYC 7R32 CPU core using the AVX2 vector extension.}
\label{fig:amd}
\end{figure}

\begin{figure}
\includegraphics[width=0.91\columnwidth]{\versusLLVMfigpath{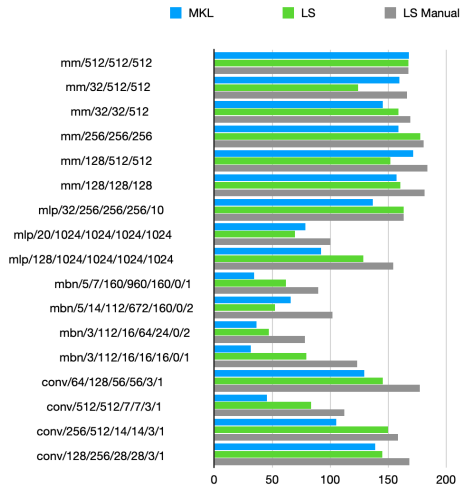}}
\caption{Comparison of MKL-DNN, Scripted, and Manual tuning using \looptool{} on an Intel(R) Xeon(R) Platinum 8124M CPU core using the AVX512 vector extension.}
\label{fig:intel}
\end{figure}

\begin{figure}
\includegraphics[width=0.91\columnwidth]{\versusLLVMfigpath{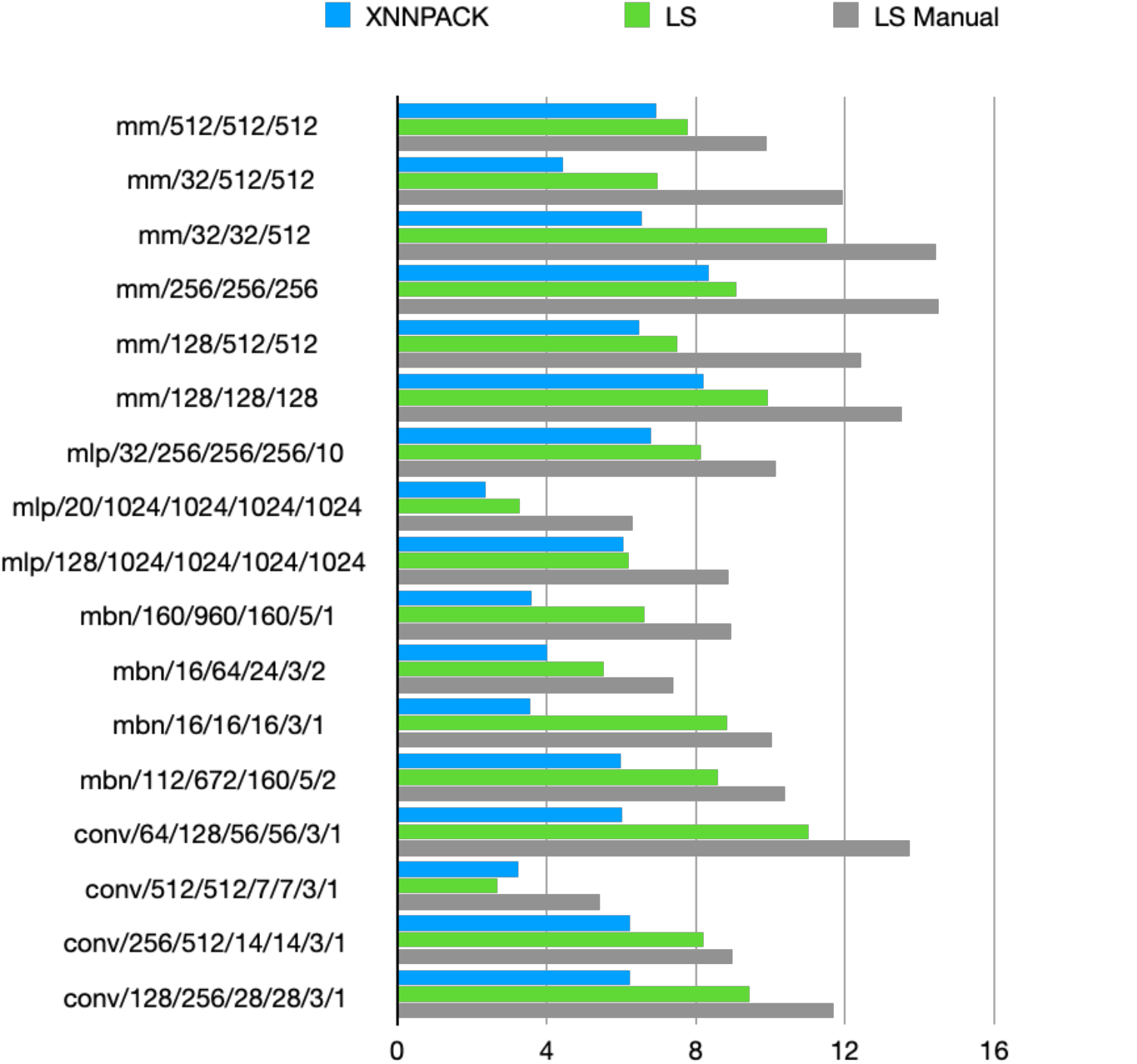}}
\caption{Comparison, in GFLOPS, of XNNPACK, Scripted, and Manual tuning using \looptool{} on an ARM Cortex A57 CPU core using the neon vector extension.}
\label{fig:a57}
\end{figure}

\begin{figure}
\includegraphics[width=0.91\columnwidth]{\versusLLVMfigpath{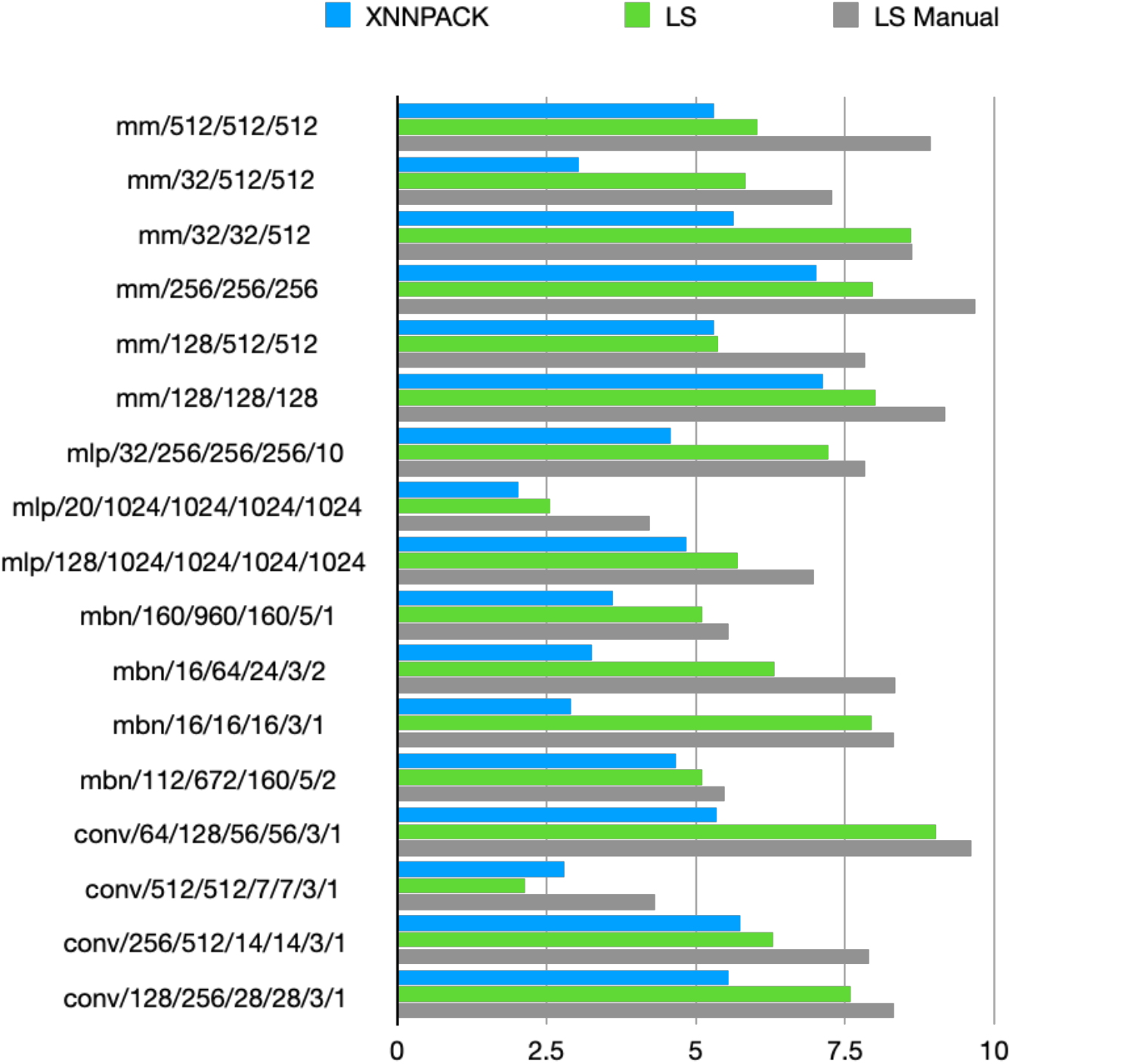}}
\caption{Comparison, in GFLOPS, of XNNPACK, Scripted, and Manual tuning using \looptool{} on an ARM Cortex A73 CPU core using the neon vector extension.}
\label{fig:a73}
\end{figure}

\begin{figure}
\includegraphics[width=0.91\columnwidth]{\versusLLVMfigpath{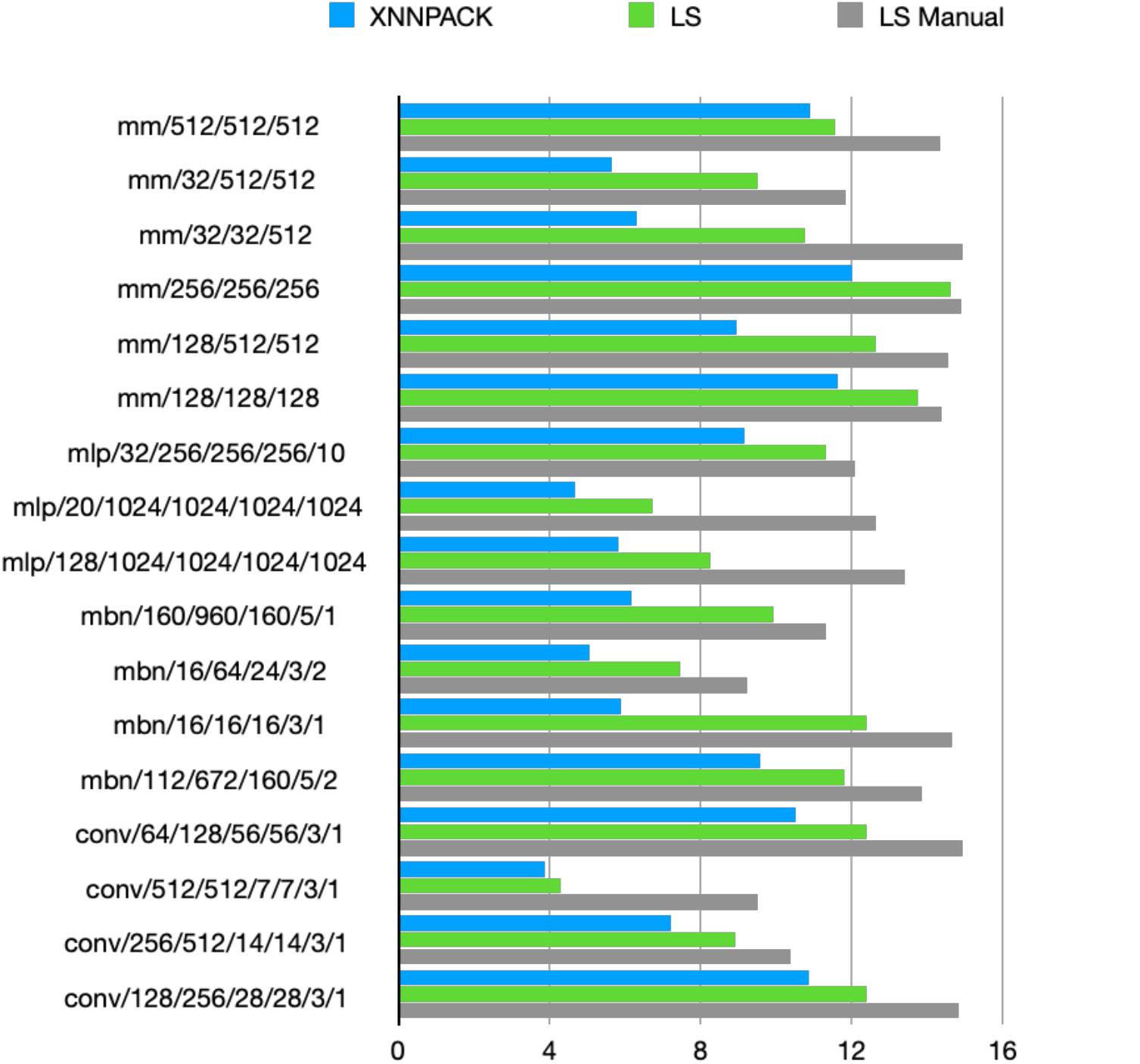}}
\caption{Comparison, in GFLOPS, of XNNPACK, Scripted, and Manual tuning using \looptool{} on an NVIDIA ARM Denver 2 CPU core using the neon vector extension.}
\label{fig:denver2}
\end{figure}

\begin{figure}
\includegraphics[width=0.91\columnwidth]{\versusLLVMfigpath{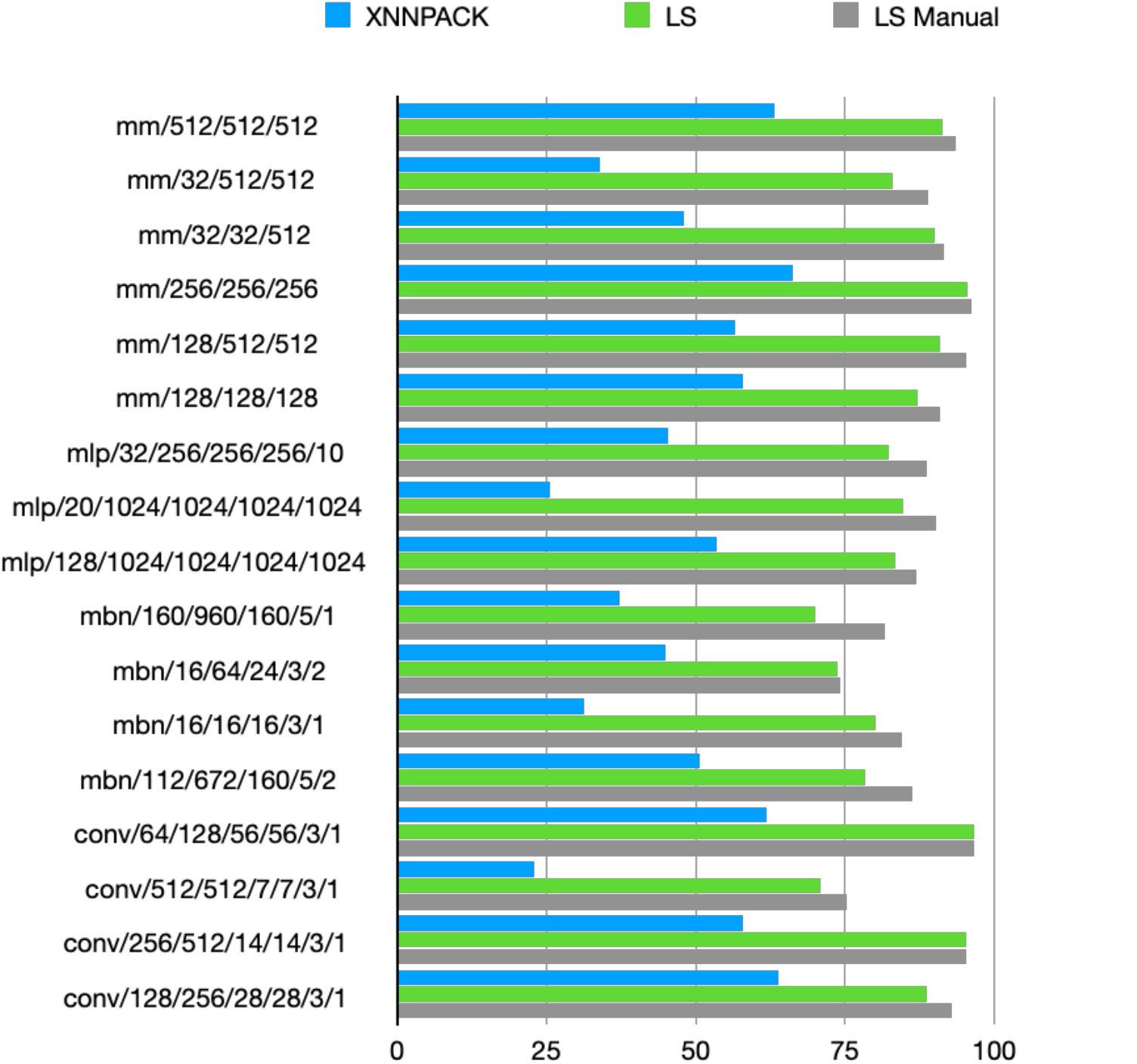}}
\caption{Comparison, in GFLOPS, of XNNPACK, Scripted, and Manual tuning using \looptool{} on an Apple M1 CPU core using the neon vector extension.}
\label{fig:m1}
\end{figure}

\subsection{Neural Network Workload: Benchmarking vs Traditional Libraries/Frameworks}

Figure~\ref{table:operator-benchmarks-runtime-tables} presents a
selection of operations common in modern neural networks.  In the
table (as well as figures
\ref{fig:amd},\ref{fig:intel},\ref{fig:a57},\ref{fig:a73},\ref{fig:denver2},\ref{fig:m1})
benchmarks prefixed with "mm" are single matrix multiplications of
varying size.  The "mlp" workloads simulate 3-layer multi-layer
perceptrons of varying batch size and hidden dimensions.  These
include ReLU activations after each fully connected layer and no final
reduction layer.  The "mobilenetV3" workloads are depth-wise separable
cells taken from the MobileNetV3 model architecture
\cite{Howard_2019_ICCV} run with a fixed spatial dimension of 12x12
and varying channel sizes.  All benchmarks are run in C++ against the
most recent version of MKL-DNN~\cite{mkl} using PyTorch \cite{pytorch}
for convenience, as well as the most recent version of
XNNPACK~\cite{xnnpack} through TFLite~\cite{abadi2016tensorflow}.

We compared the performance of the best schedules generated by
\loopstack{} using scripted tuning as described in Section 6.3.
Additionally we compared performances of manually optimized schedules
using the manual tuning TUI described in Section 6.2.  The tuning was
done by the authors by starting from the best schedules generated with
the scripted tuning, and spending no more than a dozen of minutes per
schedule.

Our results show that, our simple tuning script, implemented in less
than 500 lines of Python code, outperforms hand--optimized primitives
in most cases; additionally our manually optimized schedules
outperform the hand--tuned primitives in {\bf all} cases.
\loopstack{}, including the scripted tuner as well as the TUI, is
implemented in fewer than $30K$ lines of code~\footnote{As measured
with the standard {\bf cloc} utility}, whereas MKL-DNN and XNNPACK
both have an order of magnitude more lines of code.

\subsection{Extending \looptool{} to New Hardware}

\begin{figure}
\includegraphics[width=0.85\columnwidth]{\versusLLVMfigpath{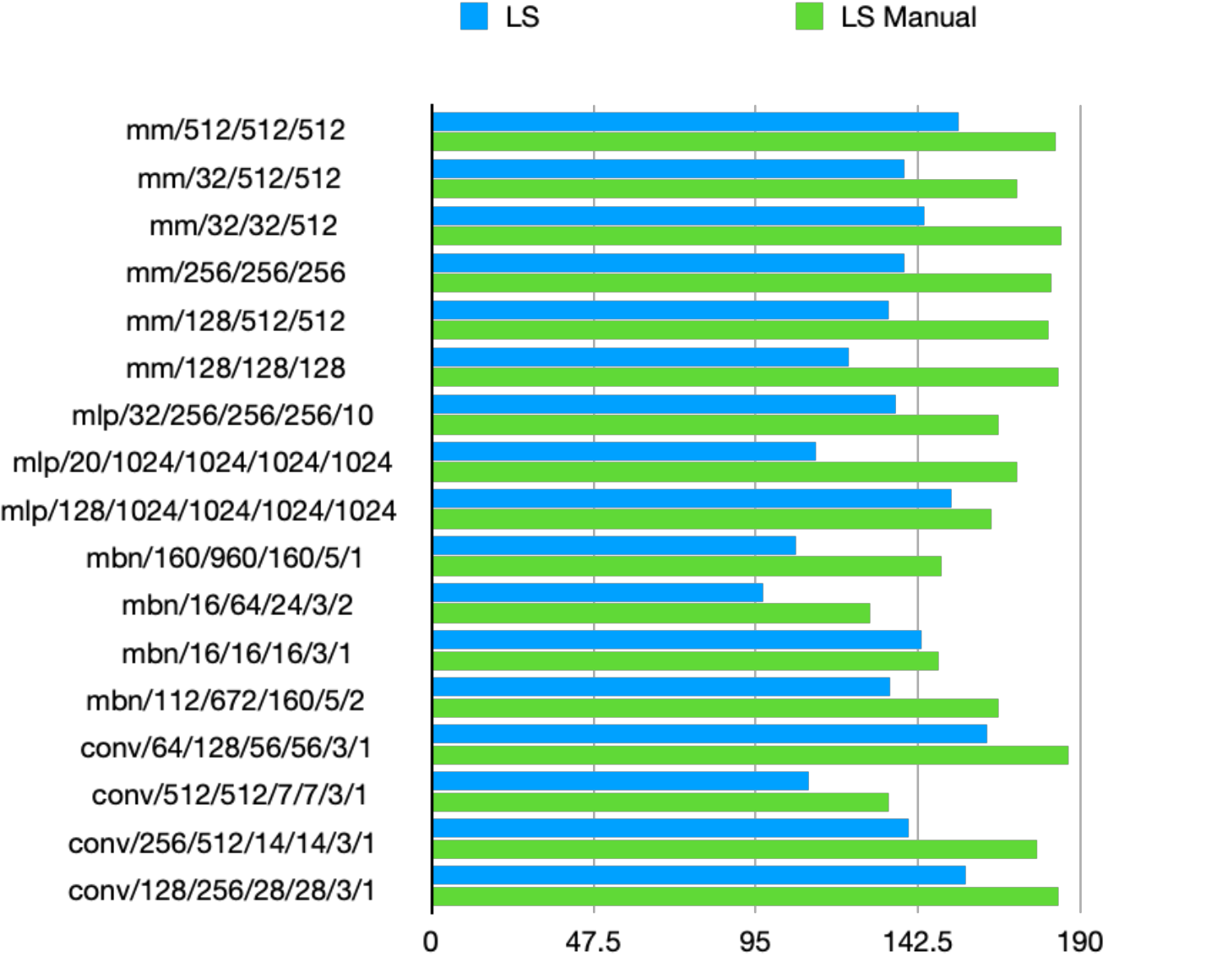}}
\caption{Results of \looptool{}'s guided search, in GFLOPS, and manual
  optimizations using the TUI on Apple's M1 chip (capable of around
  210 GFLOPS) using neon\_fp16 extension.}
\label{fig:fp16}
\end{figure}

Additionally, we perform a set of benchmarks on Apple's new M1 chip
using the neon\_fp16 extension.  Extending \loopstack{} to support
fp16 computation required a set of changes in \loopnest{} compiler
that are specific to the new target.  This changes took less than 10
engineering-days, in in fewer than 1000 lines of code.  In
Figure~\ref{fig:fp16} we show the results of scripted and manual
tuning on Apple's M1 chip, which is capable of around 210 GFLOPS peak
performances.  Our results suggest, that with a minimal amount of
expert work, as compared to traditional, hand--optimized libraries, we
can achieve extremely high utilization of new target hardware.

\section{Related Work}
Several alternative approaches have been proposed to efficiently
execute tensor based computation on hardware. They fall in one of two
broad categories: systems for high-performance tensor operations and
scheduling of tensor operations.

\subsection*{High Performance Tensor Operations}
Intel's Math Kernel Library (MKL)~\cite{mkl} provides high performance
kernels for common operations such as matrix-matrix multiplications,
matrix-vector multiplications, matrix decompositions, and
transformations. OneDNN (previously known as MKL-DNN)~\cite{onednn}
similarly aims to provide high performance implementations of common
neural network operators such as convolution, softmax, normalization,
and others, which can be composed to implement performant
pipelines. In contrast to these libraries, \loopnest{} does not
maintain separate implementations for an enumeration of different
operators and sizes, rather it provides an efficient code generation
approach to support operations over arbitrary sizes/shapes. In doing
so, \loopnest{} supports generating code for entire pipelines, rather
than composing individual operators, which can be efficient
independently but may otherwise result in subpar performance when
composed.

Tensor-Contraction Code Generator (TCCG)~\cite{tccg} is a approach for
performing tensor multiplication. To obtain high performance, TCCG
draws on three separate approaches, and integrates them into a single
tool that generates C++ code. It frames tensor multiplication as
nested loops, transposition followed by use of an efficient GEMM
implementation, and loops over GEMMs. In contrast to TCCG, \loopnest{}
is not designed specifically for tensor-tensor multiplication (though
it supports it), but rather provides support for additional operations
and targets whole-pipeline code generation. While TCCG generates C++
code (which must then be compiled), \loopnest{} uses compilation to
target x86 and aarch64. Finaly, TCCG folds in the search for a
schedule, in that it generates varied implementations (using the
different techniques underlying the tool), scores them with a
performance model, and then compiles the most promising ones.

Polly~\cite{grosser2011polly} is a a polyhedral optimizer for LLVM
code. Polly provides support for high-performance generalized matrix
multiplication, which like \loopnest{} allows users to express
computation with different multiplication and reduction
operators. However, in contrast to \loopnest{}, Polly is not a
standalone code generator and instead is designed to work as an
optimization pass in LLVM. As we have shown in our evaluation,
compiling code with LLVM is orders of magnitude slower than compiling
using \loopnest{}.


Halide~\cite{ragan2013halide} and TVM~\cite{tvm} are compilers for
general computational pipelines (with operators commonly used in image
and tensor processing) and deep learning, respectively. Both allow
users to implement high performance machine learning pipelines by
using a declarative language to express tensor computations and a
separate language for scheduling the execution of those
computations. The scheduling languages support important primitives
such as splitting, re-ordering, vectorizing, unrolling, or
parallelizing loops, among other scheduling operations. After a
computation has been scheduled, Halide and TVM use LLVM to generate
executable CPU code. As a front-end, \looptool{}'s use of two DSLs
emulates that of Halide and TVM, and provides support for a subset of
the scheduling operations. As an efficient compiler, \loopnest{}
provides the complementary goal of efficient code generation. In
particular, a goal of \loopnest{} is to become a viable CPU back-end
alternative for systems such as Halide and TVM, enabling reduced
compile times.

\subsection*{Schedule Exploration}
Several techniques have been proposed~\cite{adams2019learning,
  value_learning} to automatically schedule Halide pipelines. These
approaches represent a schedule as a collection of loop nests, which
are recursively optimized from the output stage back to the input. At
each step, they consider where to insert the storage and compute
associated with that stage and they explore different tilings of the
stage. To efficiently perform this exploration, they use search
algorithms such as beam search, guided by a performance model. The
performance model is a deep learning network trained to predict the
runtime of a pipeline given a manually defined set of features
describing the computation graph as well as the schedule.

AutoTVM~\cite{autotvm} uses simulated annealing to produce candidate
schedules, which are obtained by applying multi-level loop tiling,
loop reordering, caching sub-results, and annotating loops for
unrolling and vectorizing. The simulated annealing procedure relies on
a cost model as a scoring function. Their cost model is instantiated
to be one of various tree-learning algorithms or a neural network,
which embeds the schedule directly and applies a linear layer to
predict runtime cost. Similarly to~\cite{adams2019learning} and
~\cite{value_learning}, AutoTVM's tree regressors rely on extracting
lower-level features from the schedule such as memory access
information.

Ansor~\cite{zheng2020ansor} uses a combination of techniques to
automatically schedule TVM programs. Ansor's leverages what they term
``program sketching'', where higher level scheduling templates are
automatically generated by iteratively applying rule-based expansions
and lower level scheduling decisions are applied as annotations. Once
a complete program is sampled (by applying random annotations to a
sketch), its schedule is evolved using evolutionary search and a
learned cost model to identify promising candidates. These promising
candidates can then be benchmarked on hardware to obtain actual
performance measurements, allowing Ansor to refine its cost model for
following iterations of the search.

In contrast to these systems, \looptool{} was designed to provide
real-time assistance to experts manually exploring
schedules. \looptool{}'s feedback (e.g. arithmetic intensity, compile
time, GFLOPs achieved) allows developers to quickly validate their
intuitions. However, while \looptool{}'s design focuses on manual
search, we believe \looptool{}'s feedback could be used by future
automated search tools. In particular, automated search tools could
(due to short compile times) evaluate many more candidates through
\looptool{} than otherwise possible. Underlying the fast
responsiveness of \looptool{} is \loopnest{}'s ability to generate
high-performance code in orders of magnitude less time than
traditional backends. In a similar direction, \loopnest{}'s simple and
fast code generation, which transparently maps the schedule provided
to code generated -- eschewing traditional approaches using an
intermediate representation and multiple compiler passes -- could
facilitate extracting more meaningful schedule features for learned
cost models directly from the compiled code.


\section{Conclusion}

We presented LoopStack, a lightweight compiler designed from the
ground up for deep learning workloads. By exclusively targeting
programs that can be expressed using a generalized version of
Einstein's notation, LoopStack avoids most of the complexity inherent
in general purpose compilers. This enables it to compile deep neural
networks several orders of magnitude faster than alternatives
approaches. This also results in a much smaller code base, and
therefore a much more lightweight binary~\footnote{\loopstack{} is
implemented in fewer than 5\% lines as compared to MKL-DNN and
XNNPACK, and its binary is 200x smaller than LLVM's}, which can be
embedded in edge devices such as IoT, sensors, and so on. Furthermore,
this clarity of purpose made it implement custom optimizations that
really helps the performance of the generated code. Indeed, LoopStack
has proven to be extremely competitive against state of the art
implementations of deep neural networks on the extensive suite of
benchmark we evaluated it on.

\looptool{} also exposes its schedule feedback through a simple API,
allowing downstream systems to consume information on a particular
schedule and update that information as incremental schedule changes
are issued. This model of interaction, paired with its fast
compile/benchmarking, opens up the future possibility of using
\looptool{} in a reinforcement learning approach, where schedule
operations are ``actions'' that update the problem ``state'' (loop
structure), and feedback provided by \looptool{} functions a
``reward'' (e.g. achieved GFLOPs per second).

\bibliography{main.bib}


\end{document}